\documentclass{article}

\usepackage{multicol}
\usepackage{amsmath}
\usepackage{arxiv}
\usepackage{color}
\usepackage[utf8]{inputenc} 
\usepackage[T1]{fontenc}    
\usepackage{hyperref}       
\usepackage{url}            
\usepackage{booktabs}       
\usepackage{amsfonts}       
\usepackage{nicefrac}       
\usepackage{microtype}      
\usepackage{lipsum}
\usepackage{amsmath}
\usepackage{algorithm} 
\usepackage[noend]{algpseudocode}

\usepackage[numbers]{natbib}
\usepackage{mathrsfs}
\usepackage[utf8]{inputenc}
\usepackage{graphicx}

\def\d{{\rm d}}
\def\W{{\rm W}}
\def\mA{{\rm A}}
\def\mB{{\rm B}}
\def\mC{{\rm C}}
\def\mD{{\rm D}}
\def\mE{{\rm E}}
\def\mF{{\rm F}}

\def\R{\mathbb{R}}
\title{Partitioned Integrators for Thermodynamic Parameterization of  Neural Networks}
\author{Benedict Leimkuhler, Charles Matthews and Tiffany Vlaar\\
School of Mathematics and Maxwell Institute for the Mathematical Sciences\\University of Edinburgh\\Edinburgh EH9 3FD\\ United Kingdom}

\begin{document} 
\maketitle

\begin{abstract}
Traditionally,  neural networks are parameterized using optimization procedures such as stochastic gradient descent, RMSProp and ADAM.   These procedures tend to drive the parameters of the network toward a local minimum.   In this article, we employ alternative ``sampling''  algorithms (referred to here as ``thermodynamic parameterization methods'') which rely on discretized stochastic differential equations for a defined target distribution on parameter space.  We show that the thermodynamic perspective already improves neural network training. Moreover, by partitioning the parameters based on natural layer structure we obtain schemes with very rapid convergence   for data sets with complicated loss landscapes.
\ \\
\ \\
We describe  easy-to-implement hybrid partitioned numerical algorithms, based on discretized stochastic differential equations, which are adapted to feed-forward neural networks, including a multi-layer Langevin algorithm,  AdLaLa (combining the adaptive Langevin and Langevin algorithms) and LOL (combining Langevin and Overdamped Langevin); we examine the convergence of these methods using numerical studies and compare their performance among themselves and in relation to standard alternatives such as stochastic gradient descent and ADAM.  We present evidence that thermodynamic parameterization methods can be (i) faster, (ii) more accurate, and (iii) more robust than standard algorithms used within machine learning frameworks.
\end{abstract}

\section{Introduction} \label{sec:introduction}
Neural networks (NNs) are an important class of complex, hierarchical models which have been used in recent years for a vast range of applications.   As impactful examples we mention the exploration of chemical structure \cite{Sc2017}, medical decision making strategies for palliative care \cite{med} and Alpha Zero which is able to master a complex challenge, e.g., learning to play Go or chess, in the span of a few days \cite{AlphaZero}. Yet there remain a number of mysteries regarding the performance of neural networks, their generality, and their ultimate reliability. An important practical challenge is that neural networks require considerable computational power for training and, in many applications, re-training. Neural networks are typically parameterized/trained using variants of (stochastic) gradient descent, where the parameters -- the weights and biases of the neural network -- are updated so that the training loss (the difference between the neural network output and the `truth') is minimized. In this article, we describe new training methods suited to neural network parameterization which are applicable in a variety of settings. In this paper we focus on classification problems and single hidden layer perceptrons, although a paper on deep networks is currently in the making. Our methods combine two basic ingredients: (i) the use of additive noise within a framework of second order stochastic dynamics, and (ii) exploitation of layer structure which induces a partitioning of the parameters of the network. The algorithms we present build directly on recent ergodicity results obtained for Langevin and Adaptive Langevin algorithms \cite{LeMaSt2015,SaLeDa2017,StSaLe2019}. 

An important performance measure of a trained neural network is its capacity to generalize from its training data to unseen (test) data. Although a neural network can perform extremely well on the data on which it was trained, the algorithm used for optimization may easily end up in a minimum which does not generalize well to unseen data, a phenomenon called overfitting. Several factors appear to influence the generalization capacity of a neural network, such as the number of parameters, initialization, learning rate, stopping criterion, activation functions, and numerical method used, and no clear consensus has been reached on how these concepts interplay with one-another. 
Zhang et al. (2017) \cite{RethinkingGeneralization} found that traditional complexity measures from statistical learning theory are incapable of explaining several features of the generalization behaviour of deep neural networks (DNNs). In particular, they demonstrated that neural networks have such a high capacity that they can memorize the training data  
and can obtain zero training error on random labels (when using an architecture that gave good generalization properties when training with real labels). Explicit regularization techniques are unable to reliably attenuate this phenomenon  \cite{RethinkingGeneralization}. Regularization, which adds a parameter norm penalty term to the loss function of neural networks, is a standard approach to prevent overfitting, but does not necessarily affect the generalization error. 

So how do we find parameterizations that generalize well? Loss landscapes of deep neural networks are known to possess many low-loss minima \cite{equivalentminima}, but not all of these minima generalize equally well and different optimizers may find different solutions \cite{neyshabur2015, against_adam, Im}. The loss landscapes of neural networks are difficult to interpret due to their high-dimensionality and non-convexity. One would expect that optimizers are likely to get stuck in isolated local minima, but this was disputed by Goodfellow et al. (2015) \cite{goodfellowvinyals}, who show that a large variety of neural networks never encounter any obstacles on their optimization path, i.e., the loss from the initial to the final optimization step typically decreases monotonically. This helps to explain the success of methods such as stochastic gradient descent (SGD) in optimizing neural networks, despite the non-convexity of the objective functions.  However, in this paper we argue that the results obtained by Goodfellow et al. (2015) \cite{goodfellowvinyals} do not hold for some common types of problems, for which SGD --as well as improved optimizers such as Adam \cite{adam}-- can be shown to fail.  This failure is likely a consequence of the more complex structure of the loss landscape of these problems. This motivates the development of more sophisticated schemes for enhanced exploration of the low loss states in these settings.
 
\subsection{Bayesian perspective on neural network training}
\label{subsec:introduction_Bayesian}
In this article, we focus on the training (parameterization) process for neural networks using ideas from statistical mechanics. Neural networks approximate a function $y=f(x)$, $f:\R^m\rightarrow \R^n$ by an abstract family of maps having a simple homogeneous form. Here consider a single hidden layer perceptron network with the structure
\begin{align*}
z_j &= \varphi \left ( \sum_{i=1}^m w^{(1)}_{ij} x_i +b^{(1)}_j \right ), \ \ \ \ \  j=1,2,\ldots, d, \\
\hat{y}_k &= \varphi_{\textrm{o}} \left ( \sum_{j=1}^d w^{(2)}_{jk} z_j +b^{(2)}_k \right ), \ \ \ k=1,2,\ldots, n.
\end{align*}
where $x\in \R^{m}$ is an input data vector,  $w^{(1)}\in \R^{d\times m}$ and $w^{(2)}\in \R^{n\times d}$ are matrices that contain the weights of the various layers, $b^{(1)}\in \R^d$ and $b^{(2)}\in \R^n$ are the biases, $z \in \R^{d}$ is the networks output after passing the input data through the first layer, and $\hat{y}\in \R^n$ is the neural network's approximation of the (for a classification problem) true labels $y$. 
The function $\varphi$ is taken to be a ReLU activation function \cite{ReLU1,ReLU2} in our experiments (for other examples of activation functions we refer to \cite{Xu}). The function $\varphi_0$ is taken to be either a sigmoid, for a binary classification problem, or a softmax, for the MNIST data set. The equations define a map $\Phi:\R^m\times \R^q \rightarrow \R^n$, where $q=m d + d  + (d+1)n$ is the dimension of the parameter space, that is we have 
\[
\hat{y} = \Phi(x,\theta),
\]
where $\theta$ contains all the parameters of the neural network. The data $\mathcal{D}$ fed into the neural network consists of pairs of input data $x$ and their labels $y$. The loss function is then determined by the difference between the neural network output $\hat{y}$ and the true labels $y$.  In our experiments we used a binary cross entropy loss function (for binary classification) or cross entropy loss function (for MNIST), but many different loss functions are available \cite{Murphy}. 

In this article, we take the Bayesian perspective, that the parameters $\theta$ are defined by the data $\mathcal{D}$ only in the sense of a probability distribution defined by Bayes' formula,
\[
\rho(\theta|\mathcal{D}) \propto \rho(\mathcal{D}|\theta) \rho_{0}(\theta),
\]
where $\rho(\mathcal{D}|\theta)$ is the likelihood, which for a cross entropy loss function takes the form $\rho(\mathcal{D}|\theta) = \text{exp}\left [\sum_i \left (y_i \text{log}[\Phi(x_i,\theta)]+(1-y_i)\text{log}[1-\Phi(x_i,\theta)] \right ) \right ]$, 
and $\rho_0(\theta)$ encodes prior knowledge of $\theta$. 
The exploration of values of $\theta$ that are consistent with Bayes' rule then becomes the outstanding challenge.  When $\rho(\theta|\mathcal{D})$ is unimodal and convex it is natural to choose $\theta$ as the mode of the target distribution by maximizing the posterior probability density, a technique referred to as ``MAP,'' for ``maximum a posteriori probability,'' but in practice this does not hold for neural networks and it then becomes a challenge to identify all relevant possible parameter values, and to compare different parameter choices in terms of their relative probabilistic weight.   This task is  referred to as {\em sampling}, and thus the Bayesian parameterization problem naturally reduces to a sampling problem for the parameters of the model. While the idea of Bayesian modelling is commonplace in all areas where statistics is used, the Bayesian perspective is usually only viewed as the starting point for optimization schemes in the setting of high dimensional neural networks, due to the vast amounts of data and parameters involved \cite{Ne1995}.  
We argue here that the sampling approach provides parameterization candidates with as great or greater efficiency than standard optimization schemes.

A single parameter vector is typically not a meaningful way to characterize a model since it fails to capture the fundamental statistical nature of the relationship between data and model.
Assuming that $\theta$ is a random variable partially constrained by the knowledge of the training set, the output $\hat{y}$ of the neural network is also a random variable with its own probability distribution directly related to that of $\theta$. We can compute the mean of the output $\hat{y}$ from the parameter distribution:
\[
\bar{y} = \int \Phi (x, \theta) \rho(\theta|\mathcal{D}) \d \theta,
\]
where $\rho$ is the normalized Bayesian density for $\theta$.  We sample parameter space by generating a sequence of discrete values of $\theta$ defined by some Markov chain $\theta_0\mapsto \theta_1\mapsto \ldots$
Then we simply approximate $\bar{y}$ by   
\[
\bar{y} \approx N^{-1} \sum_{i=1}^N \Phi(x, \theta_i).
\]
Depending on the application, it is often enough to perform a draw of a singleton from the distribution of parameters thus generated. Note that the Bayesian approach gives access to the mean $\bar{y}$, as well as other statistics (such as  variances) by a similar procedure.  It is also possible to rely on the mode (or several modes, in complicated systems) as proposed parameter values and to examine the sensitivities of those parameters using averaging methods.

Several issues are raised by the use of MCMC methods, such as equilibration of the Markov process (the ``burn in'' phase, in the language of statistics), the problem of high correlation among the samples taken along the sampling path, and the actual computational procedure by which such samples can be generated efficiently.    We will not address all the issues here, but we will show that taking a sampling perspective can cast new light on some challenging problems in machine learning.    

There is a well known link between posterior sampling and MAP estimation.  Introduce the negative log posterior $l(\theta) = -\ln \rho(\theta|\mathcal{D})$, and define
\[
\rho_{\tau}(\theta) = \exp(-\tau^{-1} l(\theta))=\rho(\theta)^{1/\tau}.
\]
For $\tau=1$ we have the posterior density.  For $\tau\rightarrow 0$ we obtain a sequence of distributions which, although globally supported, have their mass confined progressively closer to the mode of the distribution.  Thus we can think of MAP as an extreme form of sampling in which the sampled distribution is more and more confined to the vicinity of the mode or modes.  In this setting, $\tau$ becomes a parameter of an embedded family of models which may be used to enhance the optimization process. An example is the process known as annealing, where $\tau$ is gradually driven from higher to lower values \cite{annealing}.\footnote{There are other ways to vary this parameter, see e.g. simulated tempering \cite{ST, ST2}
, where it is allowed to increase or decrease.} The parameter $\tau$ plays precisely the same role as temperature in statistical physics, thus the use of the term {\em thermodynamic parameterization} to describe methods that rely on this embedding (and the sampling of the associated family of probability distributions) to enhance the parameterization procedure.

We note that the full exploration of the parameter space taken as a region of Euclidean space would be implausible in high dimensions.  Neural networks are sometimes used with millions or billions of degrees of freedom and there is no conceivable way to fully explore such a space.  On the other hand a very small range of parameter values are likely to be interest (the ones that have relatively large statistical weight with respect to the probability distribution).  Moreover, there is often much to be gained by exploring parameters in the vicinity of a local maximum, i.e. by short sampling paths.   It is important to recognize that MAP estimation, as normally practiced, is local, not global, optimization.   Molecular dynamics \cite{LeMa2015}   provides an obvious illustration of the potential value of the sampling paradigm in very high dimensions.

\subsection{The parameterization process using stochastic gradients} \label{subsec:introduction_parameterization}
In this subsection, we outline the standard training procedure based on stochastic gradient descent. In subsection \ref{subsec:introduction_sdes} we shall discuss the alternative stochastic gradient Langevin dynamics method as an illustration of a sampling method.

The starting point for most training schemes is a system of ordinary differential equations of the form 
\begin{equation}
{\rm d}\theta = G(\theta) {\rm d} t, \label{eq:gradient_flow}
\end{equation}
where the function ${G}$ is the negative gradient of loss function $L(\theta|\mathcal{D})$ defined in terms of the entire training data set $\mathcal{D}$.  Such gradient systems have the feature that, along their solutions $\theta(t)$, we have
\[
\dot{L} = -\| G\|^2,
\]
implying that the loss decreases monotonically along solutions.  Since local minima are stationary points one hopes that this dynamics steadily drives the parameters to such local minima.    The most common numerical method used for solving the system is the explicit Euler method
\[
\theta_{n+1} = \theta_n + hG(\theta_n),
\]
for a choice of discretization stepsize $h>0$.
When the gradient is approximated by evaluating it on a randomly sub-sampled  partial data set we introduce gradient noise into the dynamics. This noise can be approximately modelled by replacing $G$ in each evaluation by $\tilde{G}(\theta) = G(\theta) + \Sigma (\theta) R$ where $\Sigma \Sigma^T$ is the noise covariance matrix and $R$ a standard normal random vector with iid components.  We can thus re-interpret the training process as being
\[
\theta_{n+1} = \theta_n + hG(\theta_n) +h\Sigma(\theta_n) R_n,
\]
which we recognize as Euler-Maruyama discretization of the It$\hat{\text{o}}$ SDE \cite{gardiner}
\begin{equation}
{\rm d} \theta = G(\theta) {\rm d} t +  \sqrt{h}\Sigma(\theta) {\rm d}W,
\label{eq:sgd_dynamics}
\end{equation}
using a stepsize $h$.   It is an odd feature of the process that the discretization stepsize appears in the right hand side of the SDE itself  \cite{LeSh2016}. The ratio of step size to batch size (the size of the sub-sampled data set) was shown by Jastrz\c ebski et al. (2017) \cite{JastrzebskiStorkey} to directly influence the type of parameterizations found by this method. 

The system (\ref{eq:sgd_dynamics}) is driven by multiplicative noise. Since the gradient noise defined by $\Sigma(\theta)$ is complicated, this system of SDEs has an unknown invariant distribution which will depend on the subsampling. However, if $h\rightarrow 0$ in (\ref{eq:sgd_dynamics}) it is clear that we arrive eventually at a local minimum of the loss.  One assumes that for a small value of the stepsize the consequence is that we arrive near such a local minimum, or, to be precise, due to the inherent degeneracy of neural networks, near to a manifold of local minima of the loss.

One might worry that the dynamics could be drawn frequently toward saddles where $\nabla G=0$. Although such points are unstable --under continual perturbation the dynamics bypasses the saddles and local minima are indeed eventually located-- there are other downsides to relying on gradient flow as the foundation for training algorithms. Namely, we can only ever count on gradient dynamics as a local minimization procedure.  It has, a priori, no mechanism for global exploration.   Introducing ad hoc mechanisms to increase exploration is prone to failure since, in high dimensions, there is no natural way to grid the parameter space.  

There exist an increasing number of methods, in the same class as SGD, which are based on accelerating the scheme described above. These include SGD with Momentum (see subsection \ref{subsec:Langevin_comparison}), RMSProp \cite{rmsprop}, AdaGrad \cite{adagrad} and Adam \cite{adam}.  Although the efficacy of these methods in large scale machine learning is an active area of research \cite{against_adam}, 
we have found that these can sometimes improve training, if carefully adjusted by choice of parameters (most important - the stepsize).  In most of the cases considered in this paper, Adam gave substantially better results than SGD.  

\subsection{SDE-based schemes in machine learning}
\label{subsec:introduction_sdes}
Stochastic Gradient Langevin dynamics (SGLD) \cite{WeTe2011}, the Unadjusted Langevin Algorithm (ULA) \cite{Roberts1996, Moulines} and Stochastic Gradient Nose Hoover Thermostat (SGNHT)  \cite{JoLe2011,Di2014} are examples of existing thermodynamic  sampling methods.  In SGLD one introduces an additional additive noise into (\ref{eq:sgd_dynamics}) resulting in
\begin{equation}
{\rm d} \theta = G(\theta) {\rm d} t + \sqrt{h} \Sigma(\theta) {\rm d}W + \sigma_{\rm A} {\rm dW}_{\rm A}.
\label{eq:sgld_dynamics}
\end{equation}
The additive noise is usually taken to have constant variance.\footnote{At this (formal) level we could of course combine the
two Wiener processes, but it is desirable to keep them separate since the first ($W$) only enters into evaluations of the force and must always be realized in conjunction with force evaluation in the description of a numerical method.}
SGLD is typically discretized using Euler-Maruyama, resulting in
\begin{equation}
\theta_{n+1} = \theta_n + hG(\theta_n) +h\Sigma(\theta_n) R_n+ \sigma_{\rm A}\sqrt{h} R_n^{\rm A}. \label{eq:sgld_discretization}
\end{equation}
At this stage, we see that for small $h$, the $\sqrt{h}$ term will strongly dominate the noise, and conclude that if we replace the constant stepsize $h$ by a decaying sequence of stepsizes $h_n\rightarrow 0$, we would, in the long term, expect to generate states from the stationary distribution, thus the claim that SGLD is a sampling method for a known distribution. The mathematical analysis of this method relies on the framework known as ``stochastic approximation" \cite{Ku2003}. 
The caveat of course is that such a rigorous procedure requires the use of small stepsizes which would be expected to slow the sampling process.  In practice a small bias is accepted in exchange for being able to more efficiently sample the target distribution, although Brosse et al. (2018) \cite{Brosse} argue that the high variance of stochastic gradients can limit the usefulness of SGLD in practice.

In this article we propose to use, as in SGLD, additive noise (which has an adjustable but fixed strength) to stabilize the invariant measure of the stochastic dynamics. In contrast to SGLD however, we rely on underdamped Langevin dynamics and apply state-of-the-art discretization methods \cite{LeMa2015, StSaLe2019}, which introduce additive noise within a framework of second order stochastic dynamics.
Additionally, we partition our algorithm based on the natural layer structure of the neural network.

For properties of the partitioned algorithms we draw on three recent works:  (i) hypoelliptic properties of Langevin dynamics numerical methods \cite{LeMaSt2015}, (ii) hypoelliptic properties for Langevin dynamics with configuration-dependent noise \cite{SaLeDa2017} and (iii) very recent work on weighted-$L^2$ hypocoercivity of Adaptive Langevin dynamics \cite{StSaLe2019}. To connect our methods with these works recall that our methods use additive noise and that in practice there will also be a second term with an unknown covariance arising  from the gradient approximation.  Such an approach is close to the systems treated in \cite{SaLeDa2017} where the SDEs take the form of a Langevin system where the friction matrix $\Gamma$ is allowed to vary with position $q$ (which in the results above corresponds to our parameters $\theta$)
\begin{align}
\d q & = p \ \d t, \nonumber \\
\d p & = G(q)\d t  - \Gamma(q)p \ \d t + \Sigma(q) \d {\rm W}.
\end{align}
Nondegeneracy of $\Sigma(q)$ is required for the results of \cite{SaLeDa2017} to hold, but we will obtain that by driving the system by additive noise of defined strength (in each momentum equation).  In the case of the method AdLaLa (described in subsection \ref{subsec:partitioned_AdLaLa}) which makes use of Adaptive Langevin (AdL) dynamics we have hypocoercivity results for AdL \cite{StSaLe2019}, which can be used to justify the method. These methods are based on position-independent noise. We conjecture that these hypocoercivity results can be extended to systems with position dependent noise.

\subsection{Improving stability of neural network parameterization using partitioned stochastic methods
}\label{subsec:introduction_stability}
In this paper we make use of the layer structure of neural networks to obtain partitioned algorithms that use a different optimizer for different parts of the network. We show for certain datasets that these schemes can significantly accelerate training. There have been a number of attempts in recent years to design better training strategies by relying on the detailed structure of neural networks.  For example the method AdaDelta \cite{Ze2012} attempted to use an adaptive procedure to vary the learning rate (integration stepsize) according to dimension.  
Singh et al. (2015) \cite{SiDeZhGoTa2015} looked at using different stepsizes to treat the weights and biases in different layers.  
Although the method developed showed improvements compared to using the same stepsize, the gains were small. An effort which may be potentially more relevant to our article is the work of Lan et al. (2019) \cite{LCA} which found that freezing the last layer (i.e., fixing the weights and biases in the last layer) results in significant performance gain.

We are not aware of an effort to use differential thermostatting among layers in the design of training algorithms.  In several of our experiments 
we found it advantageous to use low temperatures (even zero temperature) in the output layer but to maintain the hidden layer weights and biases at slightly elevated values.  This means that those inner parameters can rapidly explore a wide range of low-loss states.  We conjecture that it is this fluidity in the hidden layer which gives the LOL and AdLaLa methods described here their improved convergence speed.   

It is well-known that local minima can be very sensitive to small changes in the choice of hyperparameters. This sensitivity has implications for the reliability and stability of training algorithms. Standard methods to improve stability  of neural networks include $L_1$ and $L_2$ regularization a la Tikhonov \cite{LASSO1, LASSO2, ridge}. 
In our experience, these methods 
cannot be relied upon to improve the test accuracy of a classifier, where the term ``test accuracy" indicates how many of the (during the training process unseen) test data points are correctly classified by the trained neural network.

 We suggest that stochastic differential equations impose a different form of regularization, since the SDEs  incorporate additive noise.    A notable ramification is that thermodynamic parameterizations appear to give rise to classifiers whose level sets are relatively smooth compared to those produced by alternative methods. Thermodynamic parameterization thus effectively controls the distribution of weights--more precisely the distribution of the conjugate momenta associated to the weights, due to the statistical mechanical property known as equipartition of energy.  
By drawing parameter states from a sufficiently rich distribution of nearby candidate states, we show that the thermodynamic schemes produce smoother classifiers, improve generalization and reduce overfitting compared to traditional optimizers.  In our studies of spiral and other data sets herein, we did not make use of any regularization method, which did not appear to affect our obtained test accuracies. 

A benefit of using the thermodynamic parameterization approach as outlined here is to reduce the dependence of the training result on the initial conditions or the details of the mechanism of training.   Unlike in conventional stochastic gradient descent and other schemes, the methods we advocate are formally ergodic, meaning that they have a unique stationary distribution and (almost all) trajectories converge to sampling paths for the same target distribution.  This provides another way in which these schemes can improve robustness.   Even if, in practice, we are unable to see the entire distribution due to computational limitations, it is desirable that the process can in principle be improved by continued exploration.

 The per-step cost of our methods is (unless otherwise noted) roughly similar to that of the other training methods such as Stochastic Gradient Descent and ADAM, assuming the major cost of a timestep is dominated by the computation of the approximate gradient.  We examine the relative performance of the different methods in detailed series of numerical experiments.  We also examine, again in numerical experiment, the key question of the variance of the results obtained by different methods, which points to the reliability and robustness of the schemes.

\section{Langevin and Adaptive Langevin schemes}
\label{sec:Langevin}
In what follows, let $L(\theta)$ represent the overall loss defined in relation to the training data set $\mathcal{D}$ (where we have suppressed the explicit dependence of the loss $L(\theta|\mathcal{D})$ on $\mathcal{D}$ for simplicity of notation).  We suppose the loss to be piecewise smooth, Lipschitz continuous, for example as obtained using mean square error or cross entropy.  We may augment the model by a (mild) quadratic regularization to ensure confinement of the parameters.

All algorithms considered here are based on gradients.  We let  $G(\theta):=-\nabla_{\theta}L$, i.e. the (full) negative gradient of the loss, and denote by $\tilde{G}$ the truncated negative gradient obtained by selecting a randomized (uniformly sampled) finite subset of the data $\tilde{D}\subset D$ at each timestep of fixed size.  In all algorithms, the stepsize (learning rate) is denoted by $h$.  The temperature parameter used in the thermodynamic algorithms is denoted by $\tau\geq0$.  $R_n$ typically represents a vector of i.i.d. standard normal random numbers drawn at timestep $n$.

In this paper we will primarily be concerned with the use of degenerate stochastic differential equations (SDEs) as the mechanism of parameterization.   We may write these in the It$\hat{\text{o}}$ formalism \cite{gardiner} as
\[
{\rm d} Z = F(Z) {\rm d} t + \Sigma(Z) {\rm d W}.
\]
The degeneracy lies in the fact that $\Sigma$ is not necessarily of full rank.  This family of SDEs includes the underdamped and overdamped forms of Langevin (Brownian) dynamics.  It also includes various thermostat methods such as Adaptive Langevin dynamics which is based on the stochastic generalization of the (determnistic) Nos\'e-Hoover thermostat.

\subsection{Langevin dynamics}\label{subsec:Langevin_Langevin}
Consider the Langevin dynamics \cite{LeMa2015}  system of SDEs: 
\begin{align}
{\rm d} \theta &= p\, {\rm d}t,\label{eq:ld1}\\
{\rm d}p &= G(\theta) \,{\rm d}t - \Gamma p  \,{\rm d}t + \Sigma \,{\rm d W}_t, \label{eq:ld2}
\end{align} 
where $\theta$ and $p$ are the position and momentum vectors respectively, $W_t$ a standard $N$-dimensional Wiener process, and $\Gamma$ and $\Sigma$ are symmetric positive definite matrices, which we shall assume to be position-independent in the remainder of this paper. In the special case where
\[
\Sigma \Sigma^T = 2\tau \Gamma,
\]
for scalar $\tau>0$ the dynamics obeys a fluctuation-dissipation theorem, and under some mild assumptions is provably ergodic (see Section \ref{sec:properties}). This ensures that solutions of the dynamics sample the distribution $\rho_\tau(\theta,p)$ where
\[
\rho_\tau(\theta,p) := \rho_\tau(\theta) \times N(p\,|\,0,\,\tau) \propto \exp[-(L(\theta)+\|p\|^2/2)/\tau].
\]
As this stationary distribution doesn't depend on the friction term $\Gamma$, a common simplification is to simply choose $\Gamma=\gamma I$ and $\Sigma=\sqrt{2\gamma\tau}I$ in \eqref{eq:ld2}. In what follows, we will make use of this convention.

\subsection{Langevin Dynamics Splitting Methods}
\label{subsec:Langevin_splitting}
A popular way of building discretization schemes for Langevin dynamics is via the use of splitting methods \cite{LeMa2015,LeMaSt2015}. Such schemes are developed by writing the vector field as an additive decomposition (a ``splitting'') into separate parts and solving for each piece in sequence. In this article we shall use a Langevin splitting into pieces denoted $A, B$ and $O$:
\begin{align}
\text{d} \left [ \begin{matrix}\theta \\ p \end{matrix} \right ] = \underset{A}{\underbrace{\left [ \begin{matrix}  p \\ 0 \end{matrix} \right ] \ \text{d} t}} + \underset{B}{\underbrace{\left [ \begin{matrix} 0 \\ G(\theta) \end{matrix} \right ] \ \text{d} t}} + \underset{O}{\underbrace{\left [ \begin{matrix} 0 \\ - \gamma p \ \text{d}t + \sqrt{2\gamma \tau}  \ \text{d W} \end{matrix} \right ] }},
\end{align}
which, when taken individually, can be solved `exactly' in its evolution of distribution \cite{LeMa2015}. Individual update maps of the splitting pieces are then given by
\begin{equation}
\begin{aligned} \label{eq:ABO}
\mathcal{U}^A_h({\theta},{p}) &= ({\theta}+h{p},{p}), \\
\mathcal{U}^B_h({\theta},{p}) &= ({\theta},{p}+h G({\theta})), \\
\mathcal{U}^O_h({q},{p}) &= ({\theta},e^{-\gamma h} {p} + \sqrt{\tau ( 1-e^{-2\gamma h})}{R}).
\end{aligned}
\end{equation}
The last expression in Eq. \eqref{eq:ABO} can be obtained by studying the Ornstein-Uhlenbeck SDE
\begin{align*}
\text{d}{p} = -\gamma {p} \ \text{d}t + \sqrt{2\gamma \tau}  \ \text{d}{W}
\end{align*}
and observing that $\text{d}(e^{\gamma t} {p}) = e^{\gamma t}(\text{d}{p}+\gamma {p} \ \text{d}t)$. Therefore, multiply both sides of the Ornstein-Uhlenbeck SDE with $e^{\gamma t}$ to obtain
\begin{align*}
\text{d}(e^{\gamma t} {p})&= e^{\gamma t} \sqrt{2\gamma \tau}  \ \text{d}{W}\ \ \\
\Rightarrow \ e^{\gamma t} {p}(t) &= {p}(0)+ \int^t_0 e^{\gamma s} \sqrt{2\gamma \tau}  \ \text{d}{W}(s) \\
\Rightarrow \ \ \ \ \ \ {p}(t) &= e^{-\gamma t} {p}(0)+ \sqrt{\tau( 1-e^{-2\gamma t})} R.
\end{align*}

We can code schemes by changing the order in which we apply the updates, with repeated letters indicating substeps (i.e. two `A's indicate that each should be a half step of size $h/2$). For example, using the update rules in Eq. \eqref{eq:ABO} the BAOAB scheme is given by
\begin{align*}
{p}_{n+1/2} &:= {p}_n + \frac{h}{2}G({\theta}_n), \\
{\theta}_{n+1/2} &:= {\theta}_n +\frac{h}{2}{p}_{n+1/2}, \\
\hat{{p}}_{n+1/2} &:= \alpha {p}_{n+1/2}+\sqrt{\tau(1-\alpha^2)}R_n,\ \ \text{where} \ \alpha = e^{-\gamma h}, \\
{\theta}_{n+1} &:= {\theta}_{n+1/2} +\frac{h}{2}\hat{{p}}_{n+1/2},\\
{p}_{n+1} &:= \hat{{p}}_{n+1/2} + \frac{h}{2}G({\theta}_{n+1}).
\end{align*}

 In the case of Langevin dynamics applied to systems with gradient noise,
 we can understand a little the interplay of stepsize and friction by reference to a simplified model in which the gradient noise is assumed to be described by a stationary Gaussian process. Taking for simplicity scalar friction and a common scalar noise amplitude we replace Eq. (\ref{eq:ld1})-(\ref{eq:ld2}) by 
 \begin{align}
{\rm d} \theta &= p\, {\rm d}t, \label{eq:ld1a}\\
{\rm d}p &= G(\theta) \,{\rm d}t  + \sqrt{h}\sigma_G  \,{\rm d}W^{G}_t- \gamma p  \,{\rm d}t + \sqrt{2\gamma\tau} \,{\rm d}W_t, \label{eq:ld2a}
\end{align} 
where the appearance of $h$ is the consequence of the same argument presented in the introduction. In the absence of gradient noise this system samples the canonical distribution for temperature $\tau$.    We next combine the noise terms to obtain
 \[
{\rm d}p = G(\theta) \,{\rm d}t  +\sqrt{2\gamma \left [ \frac{h\sigma_G^2}{2\gamma}+ \tau\right]}  \,{\rm d W}^{C}_t- \gamma p  \,{\rm d}t.
\]
This corresponds to Langevin dynamics at the effective temperature
\[
\tau_{\rm eff} =  \frac{h\sigma_G^2}{2\gamma}+ \tau.
\]
This relation suggests to take the stepsize in proportion to $\gamma$ in order to maintain an approximately constant temperature as either parameter is varied.

\subsection{Role of Temperature}\label{roleoftemp}
\begin{figure}[htp]
\centering

\includegraphics[width=2.0in]{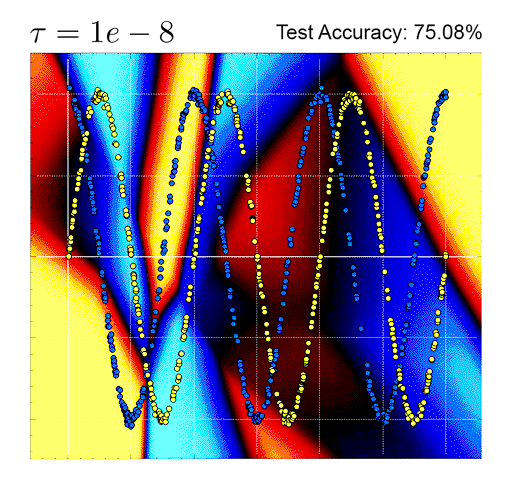}
\includegraphics[width=2.0in]{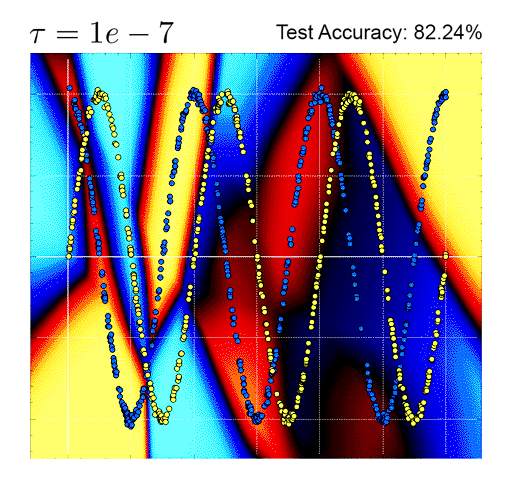}
\includegraphics[height=2.0in]{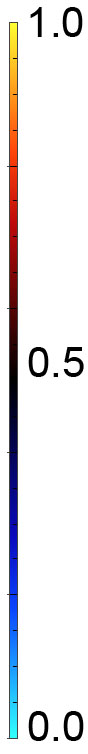}\\
\includegraphics[width=2.0in]{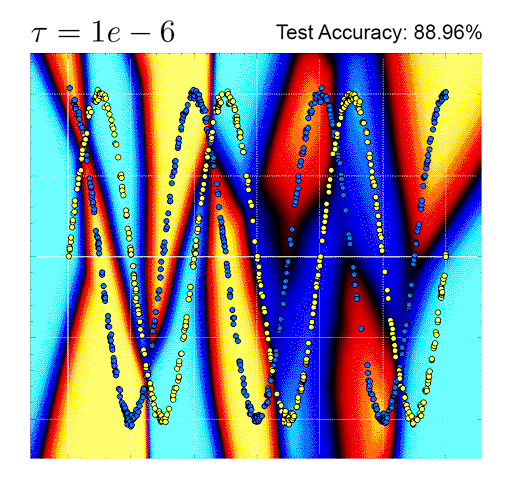}
\includegraphics[width=2.0in]{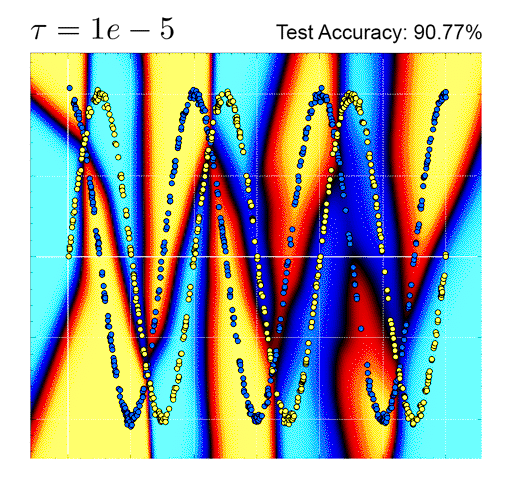}
\includegraphics[height=2.0in]{Figures/scale_f1.jpg}
\caption{The figure shows classifiers computed using the BAOAB Langevin dynamics integrator. Visually, good classification is obtained if the contrast is high between the color of plotted data and the color of the classifier, thus indicating a clear separation of the two sets of labelled data points. The same stepsize ($h=0.4$) and total number of steps $N=50,000$ was used in each training run.   The friction was also held fixed at $\gamma=10$.
A 500 node SHLP was used
with ReLU activation, sigmoidal output and a standard cross entropy loss function.  The temperatures were  set to $\tau=$1e-8 (upper left), $\tau=$1e-7 (upper right), $\tau =$1e-6 (lower left) and $\tau=$1e-5 (lower right).  The figures show that the classifier substantially improves as the temperature is raised.  The test accuracies for each run are also shown at the top of each figure. The data is given by Eq. \eqref{trigeqn} with a = 3, b = 2 and c = 0.02. We used 1000 training, 1000 test data points and 2\% subsampling.}\label{fig:baoab_trig}
\end{figure}

To make clear the role of temperature in parameterization of neural networks, we present in Fig. \ref{fig:baoab_trig} four classifiers for planar trigonometric data (see Sec. \ref{sec:models} for a full description of this data set). Each classifier was obtained using Langevin dynamics and a single hidden-layer perceptron (SHLP) for a fixed amount of work, but was parameterized with different temperatures.  Both the test accuracy and qualitative features of the classifier change with the temperature parameter, with results significantly improving as temperature increases. In further experiments we observed that further increases of the $\tau$ parameter can negatively affect the results, suggesting a `Goldilocks' temperature region of optimal efficiency.

A hypothetical model for the cause of the performance gain due to elevated temperature might be found by considering molecular diffusion on a rough energy landscape \cite{Zw1988,Po2008}. In a corrugated energy surface and at zero temperature the system will likely get stuck in local minima, lacking the required energy to overcome barriers blocking movement between states. Increasing the temperature allows weak interaction with a heat bath, randomly introducing energetic fluctuations into the system that can move it over barriers and away from local minima.   The size of the fluctuations can be related to the temperature parameter: too small and it will take a long time to cross barriers, whereas too large and the system will not be drawn towards the global minimum.

\subsection{Relation between Langevin sampling methods and certain schemes in the literature}\label{subsec:Langevin_comparison}

If we assume that a fluctuation-dissipation relationship holds and use $\Gamma = \gamma I$ then, by applying the corresponding mappings, the OBA Langevin scheme can be written as Algorithm \ref{OBA}.

    \begin{algorithm}[htp]
        \caption{The OBA Splitting Scheme}\label{OBA}
        \begin{algorithmic}[1]
            \Procedure{OBA}{$\theta$,$p$,$\gamma$,$\tau$,$T$,$h$}  
            \For{$t\gets 1$ to  $T$}   
                \State $R \gets \mathcal{N}(0,1)$\Comment{$R$ is a vector of  i.i.d. Gaussian random numbers}
                \State $p\gets \exp(-\gamma h)p + \sqrt{\tau(1-\exp(-2\gamma h))} R$
                \State $p\gets p + h \tilde{G}(\theta)$
                \State $\theta\gets\theta+hp$
            \EndFor
            \State \textbf{return} $\theta,p$ 
            \EndProcedure
        \end{algorithmic}
    \end{algorithm}
We use $\tilde{G}$ to denote the truncated negative gradient obtained by selecting a randomized finite subset of the data at each time step. We can define a family of schemes through specific choices of friction $\gamma$ and temperature $\tau$. Some schemes correspond to existing schemes in the literature. For example, setting $\tau=0$ and using finite friction we arrive at a reparameterization of the SGD with momentum scheme, with two variants given in Algorithm \ref{OBA2}. Choosing the damping parameter $\mu=h\exp(-\gamma h)$ and learning rate $\delta t=h^2$ in type I we recover the OBA scheme with $\tau=0$. Similarly in type II we reparameterize $\mu = (1+h\exp(\gamma h))^{-1}$ and $\delta t = h+\exp(-\gamma h)$. It is clear that we recover the traditional SGD scheme if $\gamma\to\infty$, or equivalently if $\mu\to0$.

    \begin{algorithm}[htp]
        \caption{The OBA Splitting Scheme with $\tau=0$}\label{OBA2}
        \begin{algorithmic}[1]
            \Procedure{OBA\_tau\_is\_zero}{$\theta$,$p$,$\gamma$,$T$,$h$}  
            \For{$t\gets 1$ to  $T$}   
                \State $p\gets \exp(-\gamma h)p + h G(\theta)$
                \State $\theta\gets\theta+hp$
            \EndFor
            \State \textbf{return} $\theta,p$ 
            \EndProcedure
            \Procedure{SGD\_with\_momentum\_I}{$\theta$,$v$,$\mu$,$T$,$\delta t$}  
            \For{$t\gets 1$ to  $T$}   
                \State $v\gets \mu v + \delta t\, \tilde{G}(\theta)$
                \State $\theta\gets\theta+v$
            \EndFor
            \State \textbf{return} $\theta,v$ 
            \EndProcedure
            \Procedure{SGD\_with\_momentum\_II}{$\theta$,$v$,$\mu$,$T$,$\delta t$}  
            \For{$t\gets 1$ to  $T$}   
                \State $v\gets \mu v + (1-\mu) \tilde{G}(\theta)$
                \State $\theta\gets\theta+\delta t\, v$
            \EndFor
            \State \textbf{return} $\theta,v$ 
            \EndProcedure
        \end{algorithmic}
    \end{algorithm}
  
     \begin{algorithm}[htp]
        \caption{The OBA Splitting Scheme with infinite friction}\label{OBA3}
        \begin{algorithmic}[1]
            \Procedure{OBA\_infinite\_friction}{$\theta$,$\tau$,$T$,$h$}  
            \For{$t\gets 1$ to  $T$}   
                \State $R \gets \mathcal{N}(0,1)$\Comment{$R$ is a vector of  i.i.d. Gaussian random numbers}
                \State $p\gets \sqrt{\tau} R + h \tilde{G}(\theta)$
                \State $\theta\gets\theta+hp$
            \EndFor
            \State \textbf{return} $\theta$ 
            \EndProcedure
            \Procedure{SGLD}{$\theta$,$\epsilon$,$T$,$\delta t$}  
            \For{$t\gets 1$ to  $T$}   
                \State $R \gets \mathcal{N}(0,1)$\Comment{$R$ is a vector of  i.i.d. Gaussian random numbers}
                \State $v\gets \epsilon \sqrt{\delta t} R + \delta t \tilde{G}(\theta)$
                \State $\theta\gets\theta+v$
            \EndFor
            \State \textbf{return} $\theta$ 
            \EndProcedure
        \end{algorithmic}
    \end{algorithm}
    
Similarly we may consider the limiting case of infinite friction and positive $\tau$ in the OBA scheme, where the momenta are redrawn from their distribution at every step. The resulting scheme (see Algorithm \ref{OBA3}) matches the SGLD scheme with a reparameterization between temperature and noise strength $\epsilon^2=\tau$ and learning rate $\delta t=h^2$. We may extend SGLD to include momentum by instead using a finite friction parameter $\gamma$ in Algorithm \ref{OBA}.
    
Thus, with a specific interpretation of the coefficients in SGD-with-momentum and SGLD we can obtain certain Langevin integrators.  All of the schemes which are of standard type are of low order of accuracy and are relatively crude in their construction; in molecular dynamics it has been shown that schemes like BAOAB substantially improve on sampling accuracy.   We thus look to the family of splitting-based methods (and further generalizations as described below) to provided enhanced training strategies.

\subsection{Adaptive Langevin and the Nos\'{e}-Hoover thermostat}\label{subsec:Langevin_AdL}
Adaptive Langevin dynamics (AdL) is a method in which the friction parameter of Langevin dynamics is automatically determined by an isokinetic control law.  The method derives from Nos\'{e}-Hoover dynamics developed by S. Nos\'{e} and W. Hoover in the early 1980s. Their proposal was to use a deterministic system to sample from the canonical ensemble \cite{No1984,Ho1985}.   
The Adaptive Langevin method, which incorporates additive noise, was first elucidated in \cite{JoLe2011} 
and has since been employed in a variety of multiscale modelling and noisy gradient settings \cite{Di2014}.  Analyses of this method can be found in \cite{LeSh2016,He2018,StSaLe2019}.

The equations take the form of a degenerate SDE system:
\begin{align}
\d \theta & = p \ \d t, \label{eq:adl1}\\
\d p & = \tilde{G}(\theta) \d t - \varepsilon \xi p \d t
+ \sigma \d \W_{\rm A}, \label{eq:adl2}\\
\d\xi & = \varepsilon \left ( p^Tp - N\tau \right ) \d t. \label{eq:adl3}
\end{align}
The hyperparameters are the coupling coefficient $\varepsilon$, the number of parameters $N$, the temperature $\tau$, and the driving noise amplitude $\sigma$.

If we assume as in subsection \ref{subsec:Langevin_splitting} that a Gaussian stationary process defines the gradient noise, then we may rewrite (\ref{eq:adl1})-(\ref{eq:adl3}) as a system with a clean gradient of the form
\begin{align*}
\d \theta & = p \ \d t, \\ 
\d p & = {G}(\theta) \d t  +\sqrt{h\sigma_G^2 + \sigma^2} \d\W_{C} - \varepsilon \xi p \d t, \\
\d\xi & = \varepsilon \left ( p^Tp - N\tau \right ) \d t. 
\end{align*}
According to \cite{JoLe2011}, this system will sample the canonical distribution at temperature $\tau$.  This implies that 
\[
\varepsilon {\rm E} \xi \equiv \gamma_{\rm eff},
\]
while
\[
h\sigma_G^2 + \sigma^2 = 2\gamma_{\rm eff} \tau,
\]
hence 
\[
\gamma_{\rm eff} = \frac{h \sigma_G^2 +\sigma^2}{2\tau}.
\]
In other words, higher additive noise $\sigma$ leads directly to larger effective friction.  Also larger gradient noise effectively increases friction.

Various discretization schemes are obtained by breaking up the AdL system into pieces (as in the discretization of Langevin dynamics) and solving the parts separately.   The maps $\mA$ and $\mB$ mentioned below are identical to those used described in the context of Langevin dynamics, although formally they need to be supplemented by an identity mapping of $\xi$. 

The simplest approach is to define the additional maps $\mC,\mD, \mE$ by
\begin{align*}
(\theta,p, \xi)\mapsto (\Theta,P, \Xi) = \mC_{h}(\theta,p,\xi)&:  \Theta:=\theta; P:= \exp(-h\xi) p; \Xi := \xi.\\
(\theta,p, \xi)\mapsto (\Theta,P, \Xi) =
\mD_{h}(\theta,p,\xi)&:  \Theta:=\theta; P:= p +\sigma \sqrt{h} R_n; \Xi := \xi.\\
(\theta,p, \xi)\mapsto (\Theta,P, \Xi) =
\mE_{h}(\theta,p,\xi)&: \Theta :=\theta; P:=p; \Xi := \xi + h \varepsilon \left [ p^T p - N\tau \right ].
\end{align*}
An obvious method 
is then defined by  the composition $\mB\mA\mC\mD\mE\mD\mC\mA\mB$:
\[
\mB_{h/2} \circ \mA_{h/2} \circ \mC_{h/2} \circ \mD_{h/2} \circ \mE_{h} \circ \mD_{h/2} \circ
\mC_{h/2} \circ \mA_{h/2} \circ \mB_{h/2}.
\]

As an alternative to the above method, one may note that the components in C and D  may be combined, resulting in an Ornstein-Uhlenbeck equation which can be analytically solved (in the weak sense).    That is, we let $\mF_h$ represent the weak solution of the equation
\[
{\rm d} p  =  - \varepsilon \xi p {\rm d} t
+ \sigma_{\rm A} \d \W,
\]
with $\xi$ held constant and substitute this F step in place of C and D. 
Care must be taken to treat small values of $\xi$ within this scheme.  We tested both methods, but did not observe notable differences in performance. We proceeded to use the first method for all our experiments.

\section{Partitioned discretization algorithms for neural networks}\label{sec:partitioned}
In layered or hierarchical models, e.g. deep neural networks, we have a natural partitioning of the parameter vector according to its role in the hierarchy.  It may be useful to treat the parameters at different levels of the hierarchy differently in the parameterization process.  In particular, it is possible that, either due to design or some feature of the network, the characteristics of the gradient noise introduced at different layer depths may differ, and it is then natural to design a method that treats the components independently.  Lan et al. (2019) \cite{LCA} observed that freezing the last layer of a neural network (while using SGD with momentum for the flexible components) can enhance the performance of training algorithms. We draw on this idea here for motivation in developing a family of partitioned algorithms that can be used to train neural networks.

In this article we focus on single hidden layer perceptrons, for which we shall use a two-part partitioning. Let $\theta = (\theta^{(1)}, \theta^{(2)})$ be a partitioning of the full parameter vector, and assume a similar partitioning of the momenta $(p^{(1)}, p^{(2)})$ as well as of the Wiener process $W(t)$.   The partitioning can be defined in various ways. For example we could group together the weights and biases at each layer 
\[
\theta^{(i)} = (w^{(i)}, b^{(i)}), \hspace{0.2in}
i=1,2. 
\]
 This is the approach we have taken in our experiments.  In extending this framework to deep neural networks one could also include several layers (or all hidden layers, say)  as one part of the partitioning.

We next describe a number of different families of partitioned integrators which could be used to take advantage of the layer structure. 

\subsection{Langevin in layers}\label{subsec:partitioned_LaLa}
The simplest idea is to use different Langevin parameters in different layers (or alternatively, Langevin with an anisotropic diagonal friction matrix).  Since temperature is purely formal in machine learning, we can, without concern for physical meanings, introduce an artificial temperature gradient by using different temperatures $\tau_1$, $\tau_2$ in the different layers. Meanwhile, we do keep the learning rate fixed at the same value across all layers and throughout training. The equations then become
\begin{align*}
\d \theta^{(i)}  & = p^{(i)} \d t,\\
\d p^{(i)} & = \tilde{G}^{(i)}(\theta) \d t - \gamma_i  p^{(i)} \d t
+ \sqrt{2\tau_i\gamma_i} \d \W^{(i)},
\end{align*}
where the indices $i = 1,2$ represent the different layers. Each subsystem can be propagated using BAOAB or some other Langevin integrator.

\subsection{Langevin-Overdamped Langevin (LOL)}
\label{subsec:partitioned_LOL}
Consider a partitioned two-part model on which we use BAOAB. 
Taking the friction to infinity in the last layer, namely taking the limit $\gamma_2 \rightarrow \infty$ results in an alternative method with a strong stabilizing property. The equations become 
\begin{alignat*}{2} 
p^{(1)}_{n+1/2} = & \ p^{(1)}_{n} + \frac{h}{2}\tilde{G}^{(1)}(\theta_n), \ \ 
&& \theta^{(1)}_{n+1} = \theta^{(1)}_{n+1/2} + \frac{h}{2} \hat{p}^{(1)}_{n+1/2},  \\
p^{(2)}_{n+1/2} = & \ p^{(2)}_{n} + \frac{h}{2}\tilde{G}^{(2)}(\theta_n), \ \ 
&& \theta^{(2)}_{n+1} =  \theta^{(2)}_{n+1/2} + \frac{h}{2} \hat{{p}}^{(2)}_{n+1/2}, \\
\theta^{(1)}_{n+1/2} = & \ \theta^{(1)}_{n} + \frac{h}{2} p^{(1)}_{n+1/2}, \ \ && p^{(1)}_{n+1} = \hat{p}^{(1)}_{n+1/2} + \frac{h}{2}\tilde{G}^{(1)}(\theta_{n+1}),  \\ 
\theta^{(2)}_{n+1/2} = & \ \theta^{(2)}_{n} + \frac{h}{2}p^{(2)}_{n+1/2},  \ \ 
&& p^{(2)}_{n+1} = \hat{{p}}^{(2)}_{n+1/2} + \frac{h}{2}\tilde{G}^{(2)}(\theta_{n+1}). \\
\hat{{p}}^{(1)}_{n+1/2} = & \ \alpha {p}^{(1)}_{n+1/2}+\sqrt{\tau_1(1-\alpha^2)}R^{(1)}_n,\ \  \text{where} \ && \alpha = e^{-\gamma_1 h}, \\
\hat{{p}}^{(2)}_{n+1/2} = & \ \sqrt{\tau_2} R^{(2)}_n,
\end{alignat*}
We refer to this method as Langevin-Overdamped Langevin or LOL. The motivation for this scheme is that it gives the possibility to increase the exploration of hidden layer structure (including the weights and biases defining the dependence on the input) while incorporating a strong dissipation in the connection to the output layer (which provides a strong stabilizing property).  In most of our experiments we also set the  temperature of the secondary partition to be zero, i.e., we set $\tau_2 = 0$. 
In this scenario one may also interpret the combined method as a sort of free energy minimization of the output layer weights and biases.

\subsection{Adaptive Langevin and Langevin in layers (AdLaLa)}\label{subsec:partitioned_AdLaLa}
As mentioned in subsection \ref{subsec:Langevin_AdL} the Adaptive Langevin method (AdL) has the property that it can automatically maintain a target temperature in a system driven by Gaussian noise.  While the gradient noise encountered in statistical approximation is not, by any means, Gaussian, it may have an important Gaussian component that can be controlled using this device (as observed in practice, see \cite{Di2014,LeSh2016}).   We therefore consider a modification of the Langevin in layers scheme in which the Adaptive Langevin method is used to manage the sampling of part of the system, thus extracting accumulated heat due to gradient noise.  

Applying Adaptive Langevin (AdL) in layers leads to the system, for $i=1,\ldots,d$:
\begin{align*}
\d \theta^{(i)}  & = p^{(i)} \d t,\\
\d p^{(i)} & = \tilde{G}^{(i)}(\theta) \d t - \varepsilon_i \xi^{(i)} p^{(i)} \d t
+ \sigma_{{\rm A},i} \d \W^{(i)}_{\rm A},\\
\d\xi^{(i)} & = \varepsilon_i \left [ \| p^{(i)}\|^2 - N_i\tau_i\right ] \d t.
\end{align*}
The parameters for layer $i$ are the coupling coefficient $\varepsilon_i$, the temperature parameter $\tau_i$, and the applied noise amplitude $\sigma_{{\rm A},i}$. Discretization then proceeds as for AdL using either of the two mentioned variants (see subsection \ref{subsec:Langevin_AdL}) or some other scheme.

It is also possible to have a partitioned algorithm with some components treated using Adaptive Langevin and others using a Langevin scheme. As a simple instance of such a method, consider the two-part ``AdLaLa" partitioning:
\begin{align}
\begin{split}
\d \theta^{(1)}  & = p^{(1)} \d t,  \\
\d p^{(1)} & = \tilde{G}^{(1)}(\theta) \d t - \varepsilon_1 \xi^{(1)} p^{(1)} \d t
+ \sigma_{{\rm A}} \d \W^{(1)}_{\rm A},\\
\d\xi^{(1)} & = \varepsilon_1 \left ( \|p^{(1)} \|^2 - N_1\tau_1 \right ) \d t,\\
\d \theta^{(2)}  & = p^{(2)} \d t,\\
\d p^{(2)} & = \tilde{G}^{(2)}(\theta) \d t - \gamma_2 p^{(2)} \d t 
+ \sqrt{2\tau_2\gamma_2}\d \W^{(2)}_{\rm A}.  
\end{split}\label{eq:AdLaLa_2layer}
\end{align}
Again we keep the learning rate fixed at the same value across all layers and throughout training. In the extreme case, where $\tau_2=0$  the second part can be viewed as a dissipated gradient system and thus we may think of this as analogous to gradient descent with momentum, but the adaptive control of the first subsystem may provide greater flexibility in the approach to the overall minimum.

\section{Model problems for classification}\label{sec:models}
We examine parameterization of fully connected single hidden-layer neural networks with ReLU activation in the context of binary classification of spiral and trigonometric data, as well as 
the MNIST data set. We found that the results were significantly different for the different problem classes, with MNIST data showing fewer substantial differences among schemes.  In order to cast some light on this issue, we use the technique of 1D linear interpolation proposed by Goodfellow et al. (2015) \cite{goodfellowvinyals} and a surface plotting technique \cite{Im}.

\begin{figure}[htp]
    \centering
    \includegraphics[width=2in]{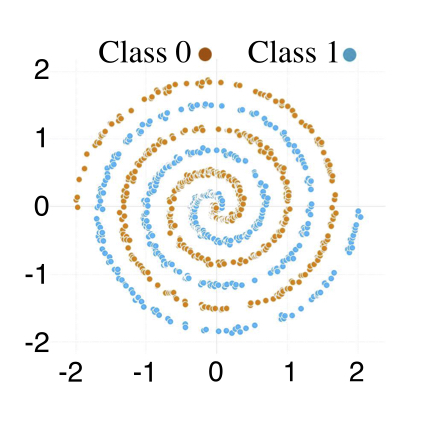}
    \hspace{0.3in}
    \includegraphics[width=2in]{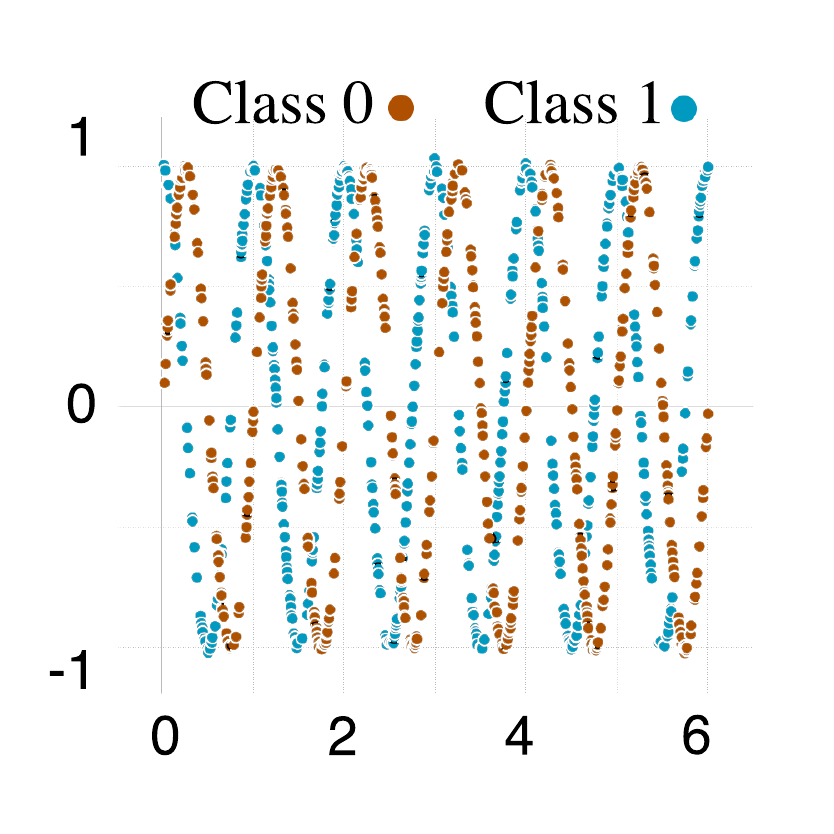}
    \caption{Spiral data and trigonometric data typical of those used in our classification studies.}
    \label{fig:class_data}
\end{figure}

The spiral data sets we use in this article are generated from the formulas
\begin{align}
 x_1& = a t^p \cos(2 b t^p \pi) +c\mathcal{N}(0,1), \nonumber \\
 x_2& = a t^p \sin(2 b t^p \pi) +c\mathcal{N}(0,1). \label{spiraleqn}
\end{align}
In these formulas, $t$ is drawn repeatedly from $\mathcal{U}(0,1)$ to generate data points, where $\mathcal{U}$ is the uniform distribution. This creates one arm of the data set, to which we assign class label 0.  The other arm constitutes a shift in the argument of the trig functions by $\pi$. We typically set $a = 2, p = 0.5$ and $c  = 0.02$, unless otherwise indicated. When we vary $b$, this directly affects the number of turns of the spiral and therefore the complexity of the problem. In the trigonometric data set the data is given for class 0 by
\begin{align}
x_1 = a t, x_2 = \cos(b t\pi) +c\mathcal{N}(0,1).\label{trigeqn}
\end{align}
Data for class 1 is generated by the same equations but with cosine replaced by sine.  Typical classification data are shown in Fig. \ref{fig:class_data}.

Depending on the choice of parameters, these can be very challenging test cases for classification, due to the consequent structure of the loss landscape.   In particular, we believe that the training algorithm  encounters significant loss-barriers for these types of data sets. For this reason, methods such as SGD and ADAM, which, up to gradient noise, monotonically decrease the loss, can easily become trapped in unsuitable states or be slowed down by the presence of saddle points.  By contrast MNIST data and related image classification problems may be relatively free of these issues. This is supported by results from Ballard et al. (2017) \cite{Ballard} who show (using molecular potential energy landscape visualization techniques) that the obtained landscape for MNIST is single-funnel-like, with only small barriers separating the different local minima from the global minimum. In contrast, they observe large barriers in the landscape for a non-linear regression problem, thus demonstrating that there exist fundamental differences in the structure of the loss landscape for different training problems. In Huang et al. (2019) \cite{ringexample} they illustrate this by designing a problem that standard optimizers will find very challenging. They set-up a binary classification problem, where they pinch the margin between two rings of datapoints, which causes any good minimizer to be ``sharp". The small volume of the corresponding basin makes these minima less likely to be found by standard optimizers. Below we include our approach to demonstrate the difference between the spiral and MNIST datasets, drawing on a method proposed by Goodfellow et al. (2015) \cite{goodfellowvinyals}.

{\em 1D Linear Interpolation:}
 Denote the initial parameter configuration by $\theta_0$ and the parameter configuration after running the optimizer by $\theta_f$. Define
\begin{align}
 \label{eq:loss-line}   \theta^*(\alpha) = (1-\alpha)\theta_0+\alpha \theta_f, \quad  \alpha \in [0,1].
\end{align}
We graph the loss $L\left (\theta^*(\alpha)\right )$ as a function of $\alpha$. At $\alpha=1$ the loss is small, while at $\alpha=0$ the loss is at a random state.

\begin{figure}[htp]
\centering
\includegraphics[width=2.2in]{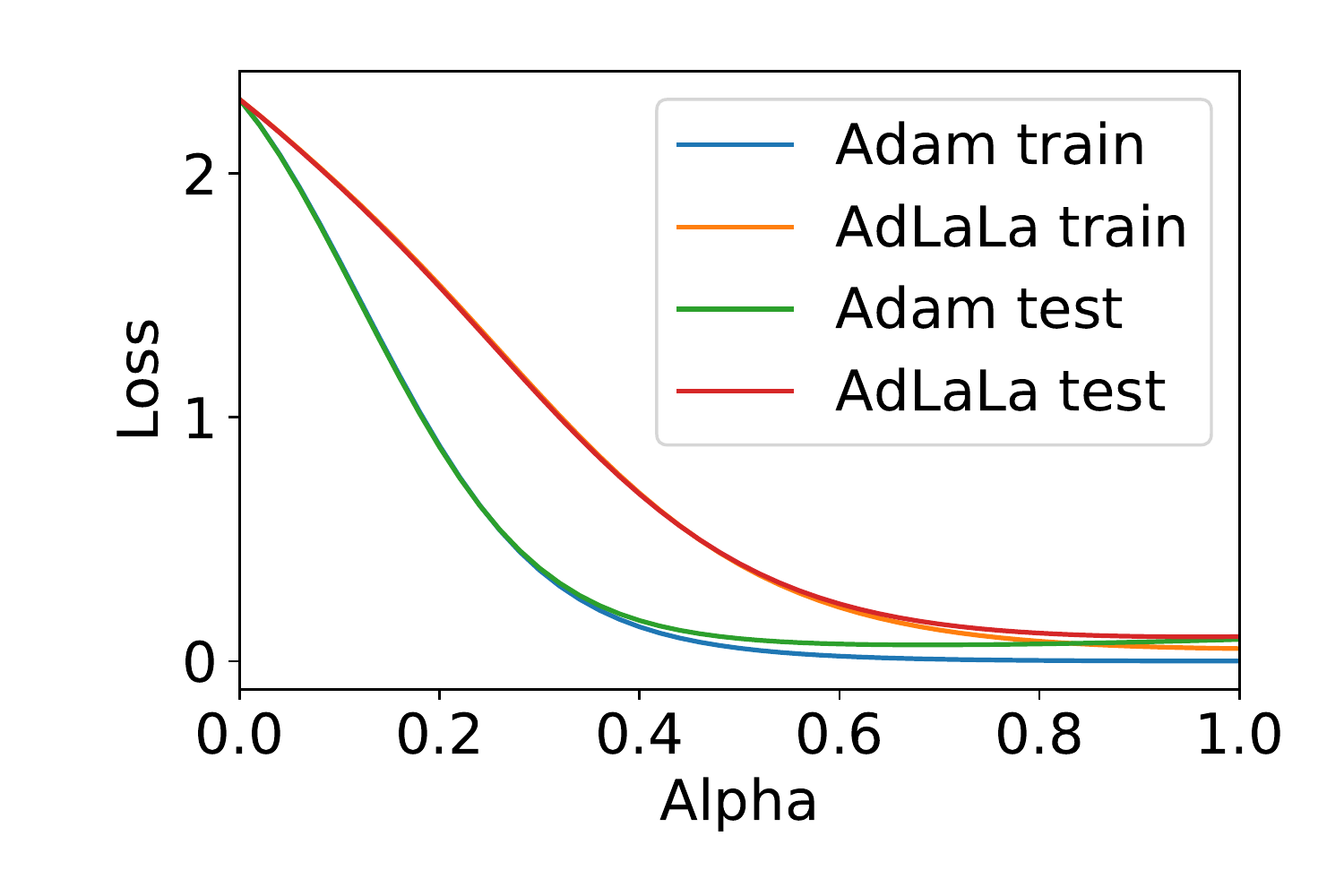} 
\includegraphics[width=2.2in]{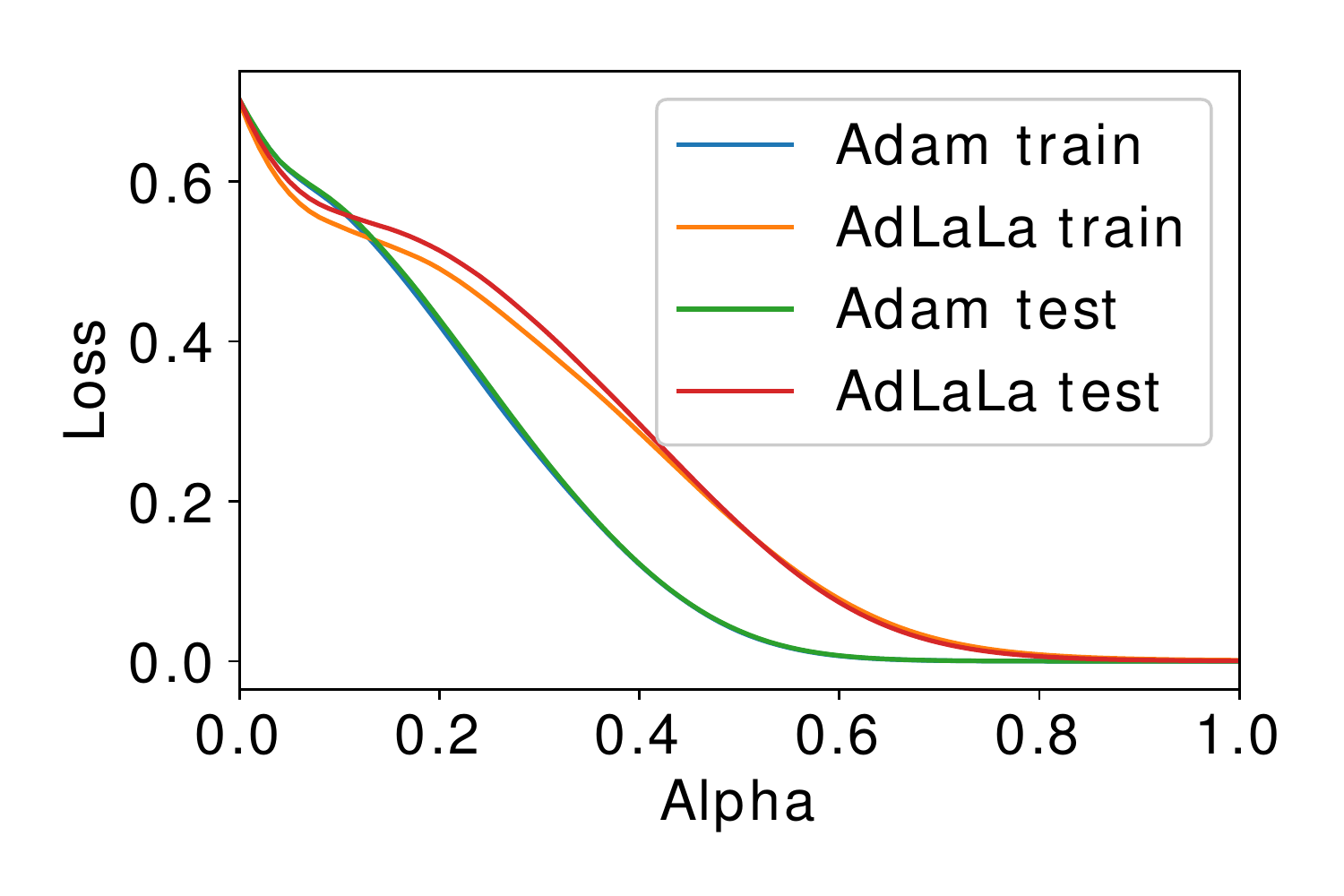}
\caption{Left: graph of the loss along the line (\ref{eq:loss-line}) for the MNIST dataset. It is clear that AdLaLa and Adam converge to different minima, although  
we used the exact same initialization for both methods. There is no evidence of a loss-barrier. Their final test loss is similar. 
Right: the same construct for a simple spiral with one turn, i.e., $b = 1$ in Eq. \eqref{spiraleqn}.  As for MNIST there is no evidence of a loss-barrier. }\label{fig:MNIST1d}
\end{figure}

For MNIST (Fig. \ref{fig:MNIST1d}) our results are
similar to the findings of Goodfellow et al. (2015) \cite{goodfellowvinyals}, specifically they observe, {``We find that the objective function has a simple, approximately
convex shape along this cross-section. In other words, if we knew the correct direction, a single
coarse line search could do a good job of training a neural network"}. 

\begin{figure}[htp]
\centering
\includegraphics[width = 2.2in]{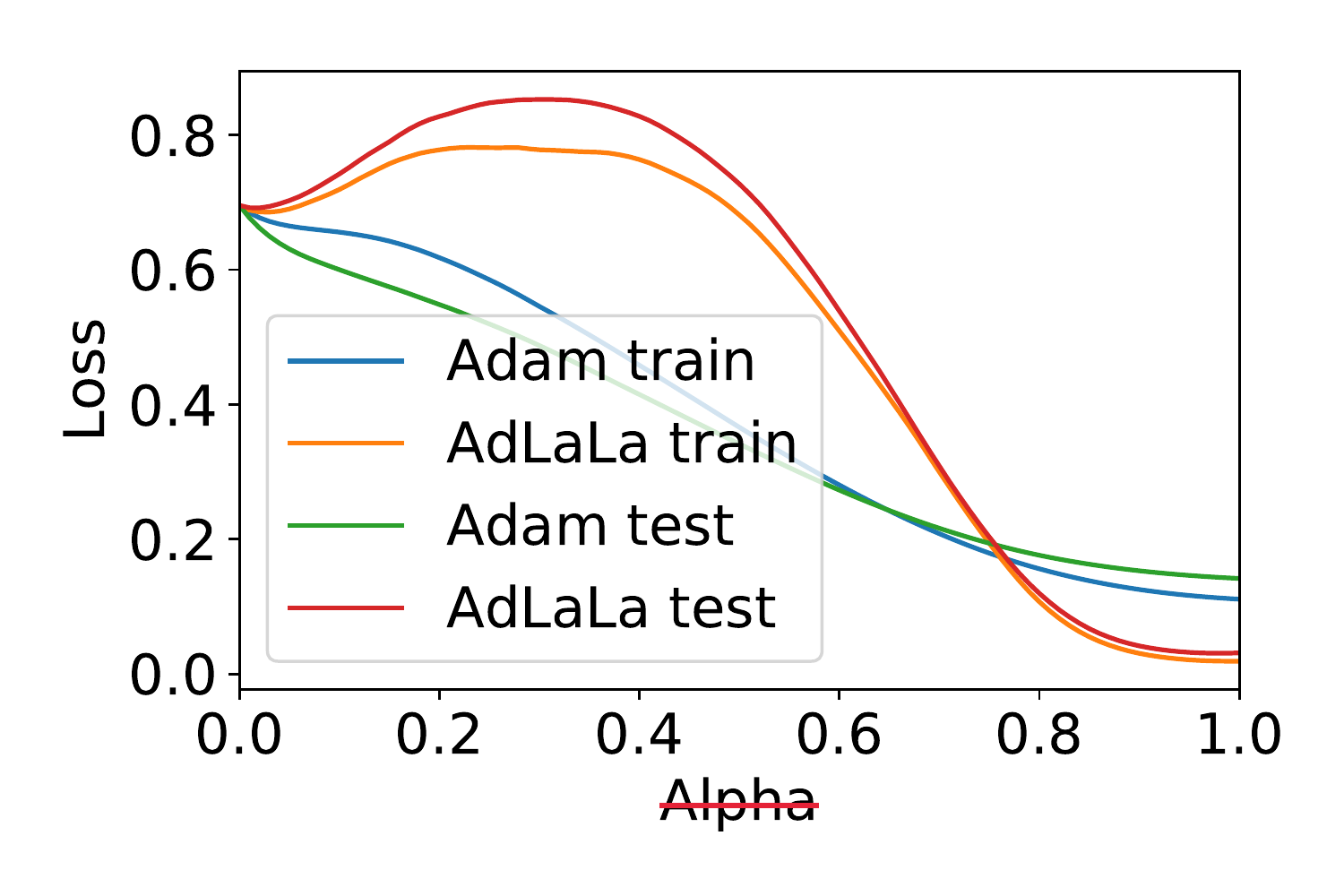}
\includegraphics[width = 2.2in]{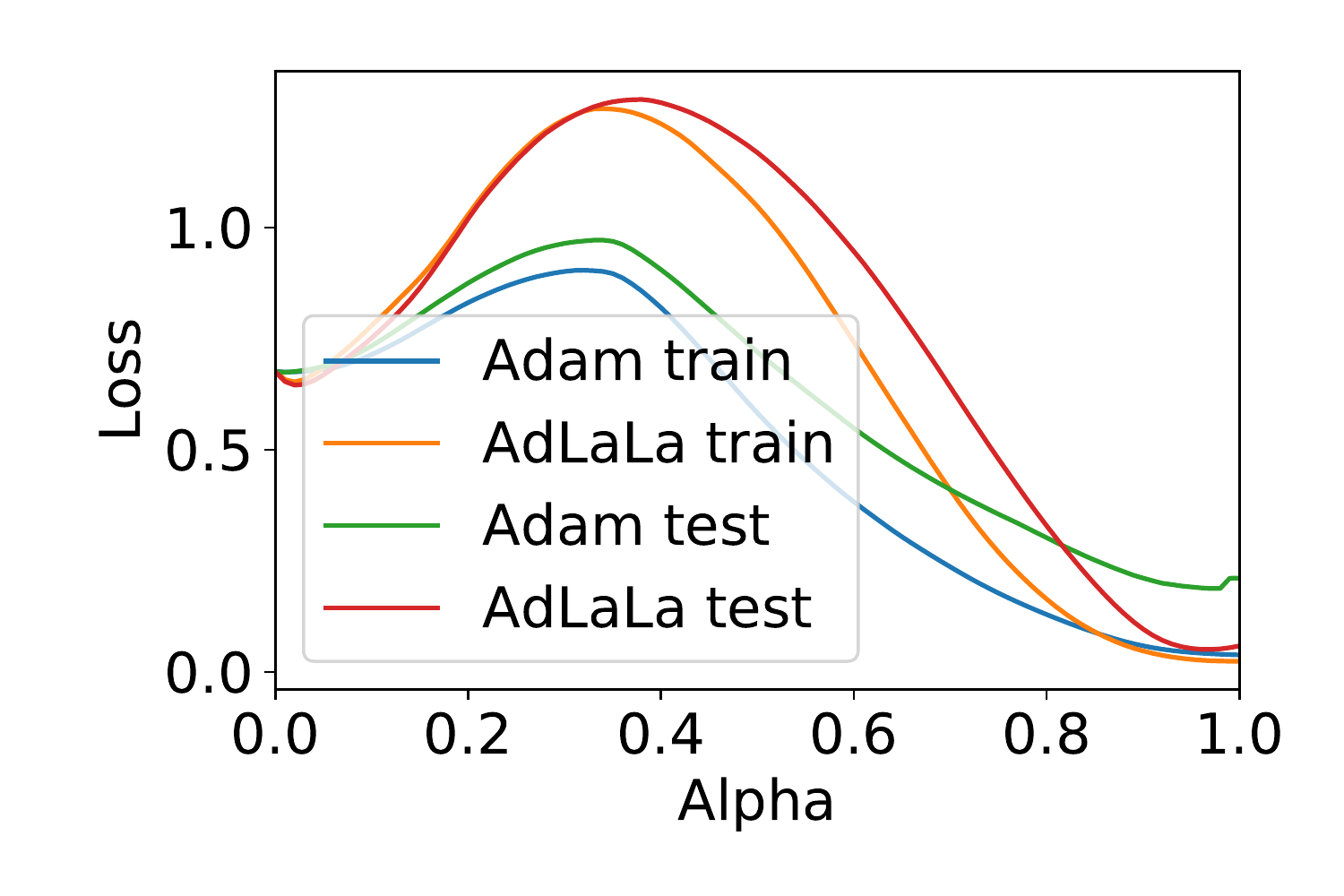}
\caption{The left and right plots are for two runs with the same parameters but different initializations. We train a 20 node SHLP on the two turn spiral dataset, i.e., $b = 2$ in Eq. \eqref{spiraleqn}, for 20,000 steps, with 500 training and test data points and 5\% subsampling.
Left: The parameterization that AdLaLa finds gives: 100\% train, 99\% test. Adam gets: 88\% train, 91\% test;
Right: AdLaLa: 100 \% train, 98 \% test. Adam: 96\% train, 94 \% test.}\label{fig:SpiralsLoss1D}
\end{figure}

 By contrast, in the spiral dataset (with more than 1 turn, i.e., $b > 1$ in Eq. \eqref{spiraleqn}) we observe that there is typically a barrier between the loss at the initial parameter configuration $\theta_0$ and the loss at the parameter configuration found by the optimizers (see Fig. \ref{fig:SpiralsLoss1D}). The barrier appears to consistently be significantly higher between $\theta_0$ and the $\theta_f$ that AdLaLa finds, than between $\theta_0$ and the $\theta_{f}$ that Adam finds ($\theta_0$ is the same for both methods). This indicates that AdLaLa finds different kinds of minima compared to Adam, which generally have lower test loss. For the trigonometric dataset, the obtained curves were generally similar to those for the spirals-2turns problem, although the height of the barrier is typically lower. We emphasize that these plots do not represent the actual path that the optimizer traverses, but do seem to point at a significant difference in the loss landscape structure of the MNIST vs. spiral/trigonometric datasets. We will elaborate on this point by constructing some surface plots.

{\em Surface plots:}
It is possible to visualize the saddle by exploring a 2-dimensional cross-section in the loss landscape.   Denote the initial parameter configuration by $\theta_0$ and now run the optimizer twice to obtain two distinct minima: $\theta_{f,1}$ and $\theta_{f,2}$.
\begin{align*}
    F_1 &= \alpha(\theta_{f,1} - \theta_0) + \theta_0, \\
    F_2 &= \alpha(\theta_{f,2} - \theta_0) + \theta_0, \\
    \theta^*(\alpha, \beta) &= \beta F_1 (\alpha)+ (1-\beta) F_2 (\alpha), \alpha \in [0,1], \beta \in [0,1].
\end{align*}
So when
\begin{itemize}
    \item $\alpha = 0$: $F_1 = \theta_0$ and $F_2 = \theta_0$. This implies that $\theta^* = \theta_0 $ if $\alpha = 0$ and $\forall \beta \in [0,1]$.  So the loss should be relatively high there, as it is the loss for a random initialization of the neural network parameters.
    \item $\alpha = 1$: $F_1 = \theta_{f,1}$ and $F_2 = \theta_{f,2}$, so the loss minima are given by $(\alpha = 1, \beta = 0)$ and $(\alpha = 1, \beta = 1$).
\end{itemize}

\begin{figure}[htp]
\centering
\includegraphics[width=2in]{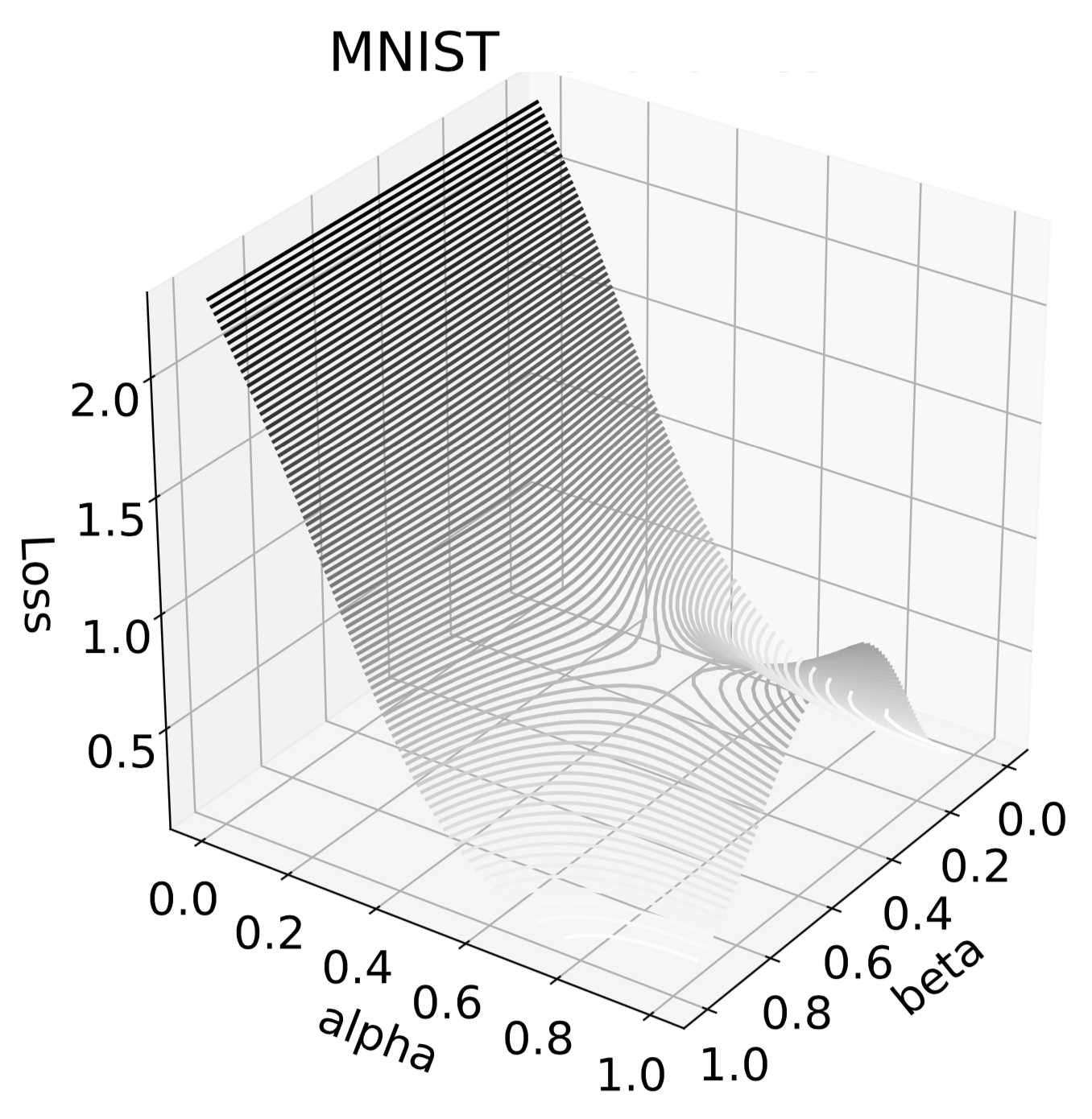}
\includegraphics[width = 2in]{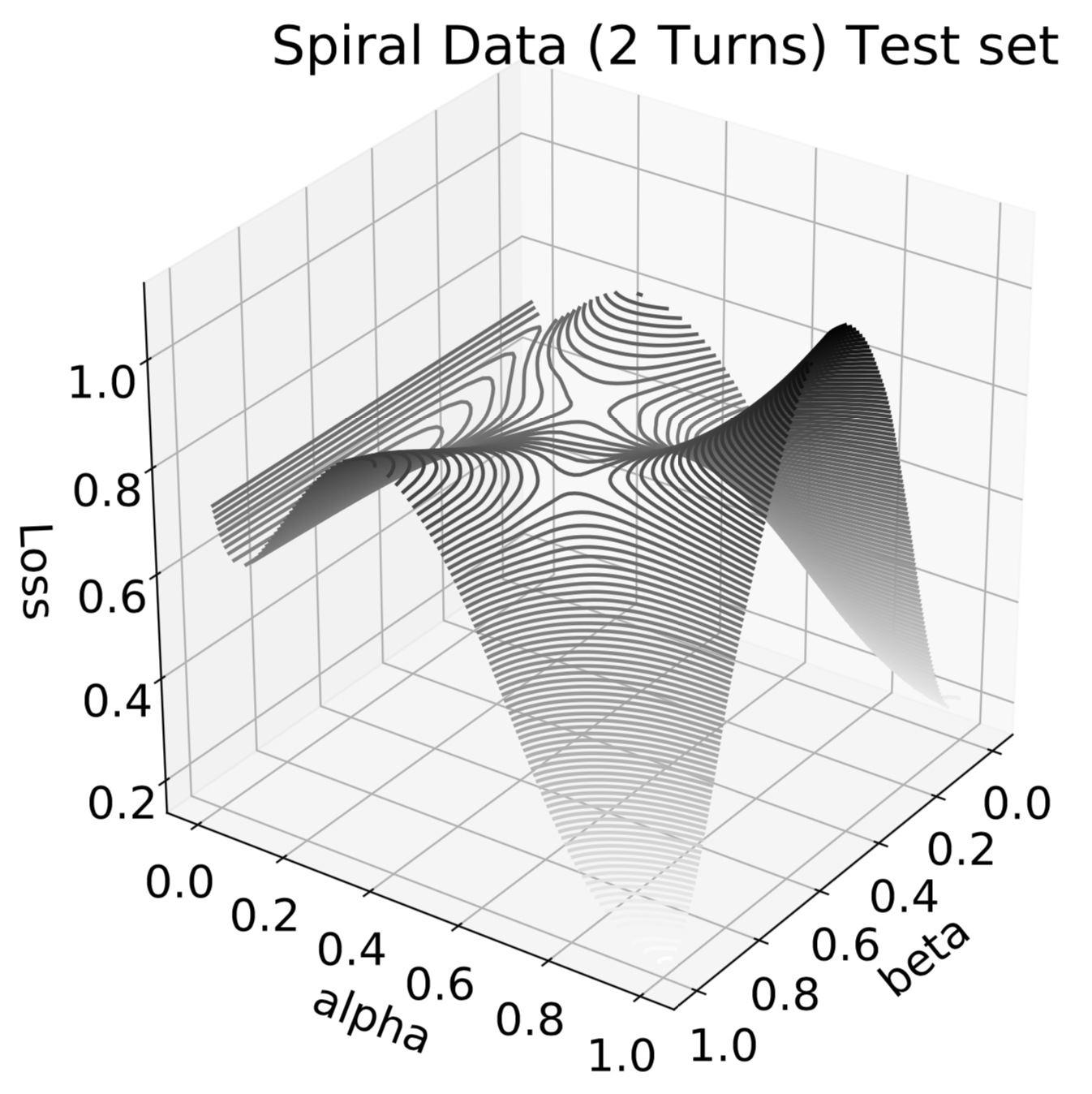}
\caption{MNIST (left) vs. Spirals (2-turn) (right) on Test.}\label{MNISTloss2d}
\end{figure}

We observe in Fig. \ref{MNISTloss2d}  that for the MNIST data set there is a consistent monotonic decline in loss along the line from the initial parameterization to the final parameterization. However, for the 2-turn spirals we frequently observe loss landscapes with saddle points in the cross-sectional plane. This seems to indicate a fundamental difference in the nature of these problems and the flexibility of the optimizers required to tackle them. We note that our low-dimensional intuitions often do not translate to the high-dimensional case: critical points with high error are exponentially likely to be saddle points, rather than local minima, which means that saddle points are thought to be the more likely cause of a possible impediment of optimization \cite{saddlepoints}.

\section{Properties of the thermodynamic parameterization methods: ergodicity, equipartition and smooth classifiers}
\label{sec:properties}

The principles of thermodynamics and the theory of hypoelliptic diffusion underpin the stochastic integrators that we have proposed previously in this article. The conditions for an SDE system to be ergodic are discussed in numerous recent works.  We summarize these as used in our own recent studies of ergodic properties of Langevin and generalized Langevin equations.

Consider the Langevin system (\ref{eq:ld1})-(\ref{eq:ld2}).  The starting point for analysis of SDEs is the Fokker-Planck equation \cite{gardiner}
\[
\frac{\partial \rho}{\partial t} = \mathcal{L}^{\dagger} \rho,
\]
where
\[
\mathcal{L}^{\dagger} \rho = -\nabla_{\theta} \cdot (p\rho) + \nabla_p \cdot \left (\left [-G(\theta) + \gamma p \right ] \rho \right )
+\gamma \tau \Delta_p \rho,
\]
where $\Delta_p$ is the Laplacian in the momenta components only
\[
\Delta_p = \sum_{i=1}^N \frac{\partial^2}{\partial p_i^2}.
\]
Assuming $G$ is smooth it is possible to find conditions which ensure that the system is ergodic in a weighted $L^{\infty}$ space; this is the usual approach based on Harris chains that one finds described in detail in the excellent book of Meyn and Tweedie \cite{MeTw1993}.  For Langevin dynamics, the analysis was first carried out in detail in \cite{MaStHi2002}.     More recently, an alternative framework has become available which is in many ways more directly suited to applications of SDEs to machine learning.   This is the method described in the work of Dolbeault et al. (2009) \cite{DoMoSc09}, which allows the derivation of exponential convergence rates when the Fokker-Planck operator is considered in a suitable subspace of $L^2(\mu_{\tau})$, i.e. weighted by the canonical invariant measure $\mu_{\tau}$. The method can be shown to give convergence estimates for underdamped Langevin dynamics.  

In very recent work, the same framework was applied to the Adaptive Langevin dynamics system \cite{StSaLe2019}.   The power of $L^2$ estimates is that they can be used to establish a Central Limit Theorem which is very important in statistical applications.

Although we have not yet looked in detail at hypocoercivity for the more complicated partitioned methods discussed here such as AdLaLa, LOL etc. (and it is certainly beyond the scope of this paper to do so), we expect that the weighted $L^2$ approach as used for AdL in \cite{StSaLe2019} could be applied to these systems as well, in order to establish the ergodic property.  By contrast, for any of the deterministic schemes mentioned and for schemes relying solely on gradient noise, ergodicity is very unlikely to hold and we are unaware of any mathematical technique that could be used for their analysis.  When additive noise is combined with gradient noise, assuming enough boundedness, a unique invariant measure still can be shown to exist using weighted $L^{\infty}$ techniques  \cite{SaLeDa2017}.

With regard to the discretized systems with additive noise, it seems likely that similar ergodic estimates can be formulated and proved.  For example in \cite{LeMaSt2015}, we have already examined in detail the ergodic properties of Langevin splitting integrators such as BAOAB on weighted $L^{\infty}$ spaces.

\subsection{Equipartition property}\label{subsec:properties_equipartition}
One of the most powerful consequences of ergodicity is {\em equipartition of energy} which simply states that the mean kinetic energy of all degrees of freedom, in thermal equilibrium, is constant.  This property can easily be derived by leveraging the uniqueness of the stationary distribution and then through direct integration of the Gibbs density, that is, for each $i$,
\[
\frac{\int_{-\infty}^{\infty}\int_{-\infty}^{\infty}\cdots \int_{-\infty}^{\infty}p_i^2 \exp\left (-\tau^{-1} \left [\sum_{i=1}^N p_i^2/2 +L(\theta) \right ] \right ) {\rm d}^N p {\rm d}^N{\theta}}
{
\int_{-\infty}^{\infty}\int_{-\infty}^{\infty}\cdots \int_{-\infty}^{\infty} \exp\left (-\tau^{-1} \left [\sum_{i=1}^N p_i^2/2 +L(\theta) \right ] \right ) {\rm d}^N p {\rm d}^N{\theta}
}=\tau.
\]
We confirmed experimentally that the magnitude of the squared momenta are approximately controlled by the set temperature value in AdLaLa and LOL.
Because of equipartition we are assured that every weight will be driven directly by a momentum coordinate which has a Gaussian distribution.  While we cannot predict the distribution of the weights themselves, since the full complement of weights are coupled intricately through the network structure, we can be sure that they will explore the full available configuration space.   Even if small, all weights should be active during training using a thermodynamic method.

\subsection{Weight Distributions}
\label{subsec:properties_weights}
\begin{figure}[htp]
\begin{center}
   \includegraphics[width=4.5in]{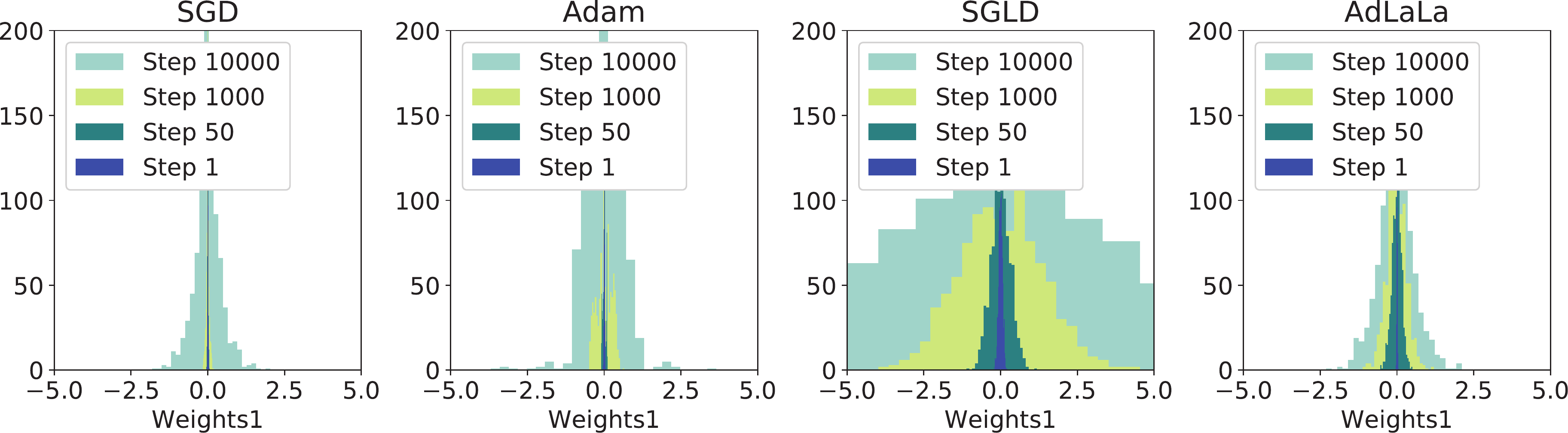}
\includegraphics[width=4.5in]{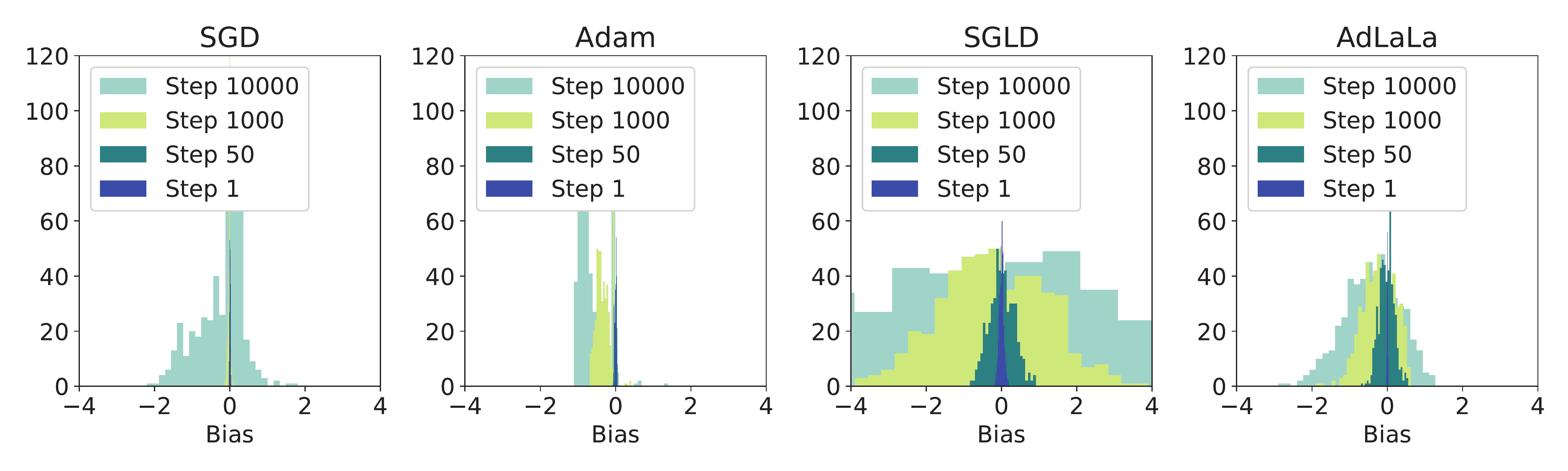}
   \caption{Weight and bias distributions for the 2-turn spirals dataset at different times and for different methods. Parameter settings: $h_{\text{SGD}} = 0.2, h_{\text{Adam}} = 0.005$, SGLD: $h_{\text{SGLD}} = 0.1$ and $\sigma_{\text{SGLD}} = 0.01$. AdLaLa: $h_{\text{AdLaLa}} = 0.25, \sigma_A = 0.01$, $\tau_1 = \tau_2 = 10^{-4}, \epsilon = 0.1$ and $\gamma = 0.5$. Test accuracy at step 50: 0.66 (SGD), 0.65 (Adam), 0.61 (SGLD), 0.62 (AdLaLa);  at step 1000: 0.66 (SGD), 0.89 (Adam), 0.68 (SGLD), 0.82 (AdLaLa); at step 10000: 0.96 (SGD), 0.99 (Adam), 0.74 (SGLD), 0.99 (AdLaLa).}
   \label{fig:weightevolution_2spiral}
   
  \end{center}
\end{figure}

\begin{figure}[htp]
\begin{center}
   \includegraphics[width=4.5in]{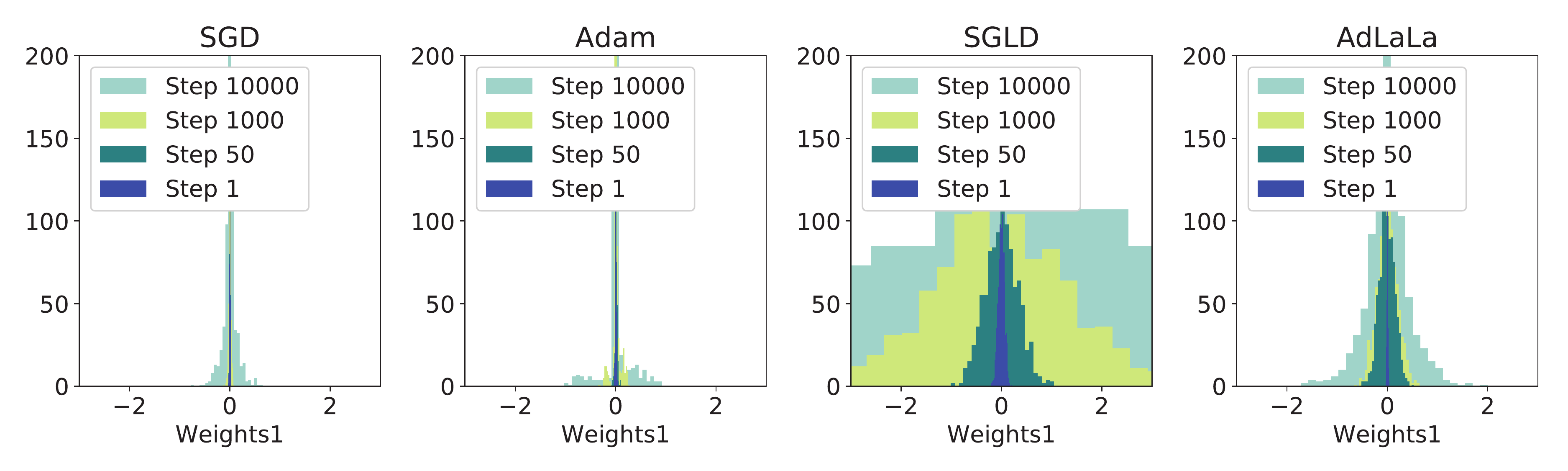}
  \caption{Evolution of weights for the 4-turn spiral problem. Same parameter settings as in Fig. \ref{fig:weightevolution_2spiral}, but $\gamma = 0.1$ in AdLaLa. Test accuracy at step 50: 0.5 (SGD), 0.58 (Adam), 0.52 (SGLD), 0.45 (AdLaLa); at step 1000: 0.56 (SGD), 0.55 (Adam), 0.5 (SGLD), 0.62 (AdLaLa); at step 10k: 0.58 (SGD), 0.67 (Adam), 0.54 (SGLD), 0.8 (AdLaLa).}\label{fig:weightevolution_4spiral}
  \end{center}
  \end{figure}

We observe fundamental differences in the parameterizations obtained by sampling methods, such as SGLD and AdLaLa, compared to standard optimizers, such as SGD and Adam. We shall illustrate this by plotting the evolution of the obtained weights and biases over time for both the spirals 2-turn data set (see Fig. \ref{fig:weightevolution_2spiral}) and the complicated spirals 4-turn data set (see Fig. \ref{fig:weightevolution_4spiral}). We use a SHLP with 500 nodes and ReLU activation, 1000 training data points and 2\% subsampling.  We distinguish between two sets of weights: those linking the input layer to the hidden layer, weights1 (first row), and those linking hidden layer to  output, weights2. We also show the distribution of biases in the hidden layer (second row). We do not show weights2, as their distribution is very similar to those of weights1. Weights2 do typically assume a larger values than weights1, but this is the same for all methods evaluated here. 

The sampling methods rapidly excite a large amount of parameters. This is clearly visible by comparing the obtained weight/bias distributions after a mere 50 steps (dark green colour in the figures) for the different methods. For the easier 2-turn spirals data set (see Fig. \ref{fig:weightevolution_2spiral}), minima are easier accessible and fewer nodes are required to obtain a good classification, which leads SGD and Adam to be able to find good minima without exciting all the weights and biases. For the complicated 4-turn spirals data set however, AdLaLa makes much faster headway towards high test accuracies (see Fig. \ref{fig:weightevolution_4spiral}), whereas Adam and SGD appear to be stuck in a parameterization with many small weights/biases. We observe that although SGLD consistently assigns much larger values to the parameters it obtains than AdLaLa, this does not appear to be beneficial for its performance on the test data set.

\begin{figure}[htp]
\begin{center}
\includegraphics[width=4.5in]{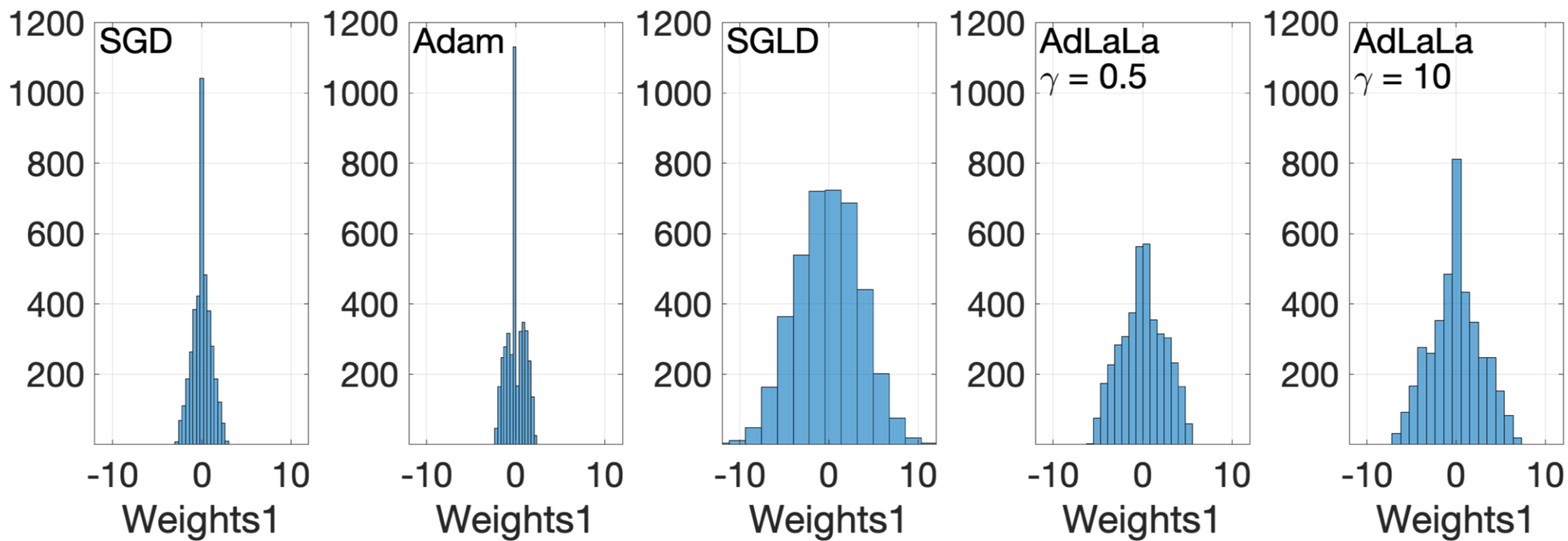}
\hspace*{0.2cm} \includegraphics[width=4.6in]{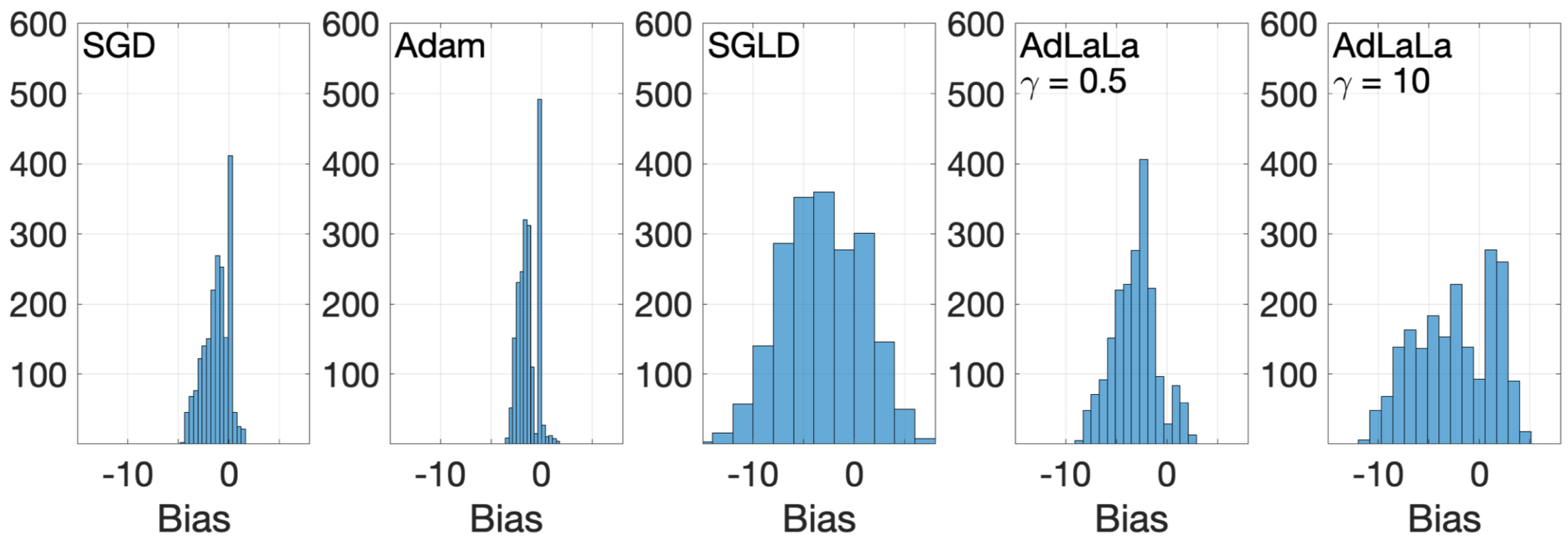}
\caption{Obtained parameter distributions over 100 runs after using different optimizers for the 2-turn spiral problem for 10K steps. Parameter settings: $h_{\text{SGD}} = 0.1, h_{\text{Adam}} = 0.005, h_{\text{SGLD}} = 0.1, \sigma_{\text{SGLD}} = 0.1$, AdLaLa has $h_{\text{AdLaLa}}= 0.25, \tau_1 = \tau_2 = 10^{-4}, \sigma_A = 0.01, \epsilon = 0.1$, $\gamma = 0.5$ (left) and $\gamma = 10$ (right). Average test accuracies: 
SGD: 79\%, Adam: 83.7\%, SGLD: 78\%, AdLaLa ($\gamma = 0.5$): 93.4\%, AdLaLa ($\gamma = 10$): 85.5\%.
}
\label{fig:weightdistributions100runs}
\end{center}
\end{figure}

As figures \ref{fig:weightevolution_2spiral} and \ref{fig:weightevolution_4spiral} only showed the obtained parameter distributions of a single run of the optimizers, we will now validate that these results are consistent over many different runs. To do so we plot all the weights obtained over 100 different runs into one histogram (see Fig. \ref{fig:weightdistributions100runs}). This shows the overall trend of the parameter distributions. To obtain these results we used a SHLP with 20 nodes, 500 training data and 5\% subsampling, for the spirals 2-turn data set. 

It is clear that SGD and Adam obtain parameterizations which have many (close to) zero weights and biases. The same was observed for different stepsizes and different batchsizes. In SGLD and AdLaLa most weights and biases appear to be equally activated.  We suggest that this is a consequence of the ergodicity and equipartition property of the latter methods.
We also note that for most optimizers their obtained layer-2 weights tend to be larger than layer-1 weights, but this changes if one increases the $\gamma$ parameter in the AdLaLa method. We note that for the LOL method (not shown in the figure) 
weights can take on both very small and very large values; in particular layer-1 weights and biases can take on values of the order $10^2$. This may indicate a possible instability and appears to be linked to larger classifier gradients.  

Out of the 100 runs we also compared the parameter distributions for the run with the worst test accuracy vs. the run with the best test accuracy. We observe that Adam performs worse if a larger percentage of the weights and biases are zero. The same holds for LOL, although the difference in accuracies is less dramatic between the worst and best run (10\% difference in test accuracy for LOL, 35\% for Adam). For AdLaLa there is even less variation in the accuracies obtained and the weights appear to be always approximately equally distributed around zero.

\section{Numerical Studies with Thermodynamic Parameterization Methods}\label{sec:numerics}

Tests of the various methods were conducted using three separate codes for cross-validation and verification of consistency:
\begin{itemize}
\item We used custom a PyTorch-based \cite{Pytorch} system  [version 1.0.0].
\item We implemented the schemes into the latest version of the DLIB package \cite{dlib09} written in C++. 
\item We created a custom native C++/QT application to perform rapid visual exploration of the training algorithms.
\end{itemize}
Code that implements the algorithms described in the article is available on 

\href{https://github.com/TiffanyVlaar/ThermodynamicParameterizationOfNNs}{github.com/TiffanyVlaar/ThermodynamicParameterizationOfNNs}.

\subsection{Choice of hyperparameters}
Despite the high accuracy and rapid convergence of our partitioned schemes (as we illustrate below), a practitioner may consider the relatively large amount of hyperparameters of these methods to be a disadvantage. We wish to emphasize that one can use certain rules of thumb to select the values of these hyperparameters, which significantly reduces the work required in tuning. Additionally, we note that our methods appear less sensitive to the choice of initialization (see subsection \ref{subsec:numerics_robustness}) or the subsampling batchsize, which can be viewed as reducing another aspect of ``tuning'' and  is thus a major performance gain compared to e.g. SGD or Adam. We also do not change the stepsize (learning rate) throughout training and are still able to obtain great performance using our methods. 

As rule of thumb for AdLaLa, one can typically set the temperatures of all layers to $10^{-4}$, the coupling coefficient $\epsilon \in [0.05,0.1]$, additive noise $\sigma_A \in [10^{-2},10^{-4}]$  and obtain good performance.   In some cases there was an advantage to using a lower temperature for the output layer.   The value of $\gamma$ (associated to the output layer) appears to be linked to the stepsize, consistent with the discussion of subsection \ref{subsec:Langevin_splitting}.  We recommend $\gamma \in [0.1,10]$, in typical cases. In some of our tests much higher values were used with good effect, whereas smaller values typically lead to stepsize restriction.  Generally, we can use stepsizes for AdLaLa  which are similar to or even larger than those for SGD or SGLD, but for  some of the harder problems the stepsize needed to be modestly reduced. For all spirals examples using a SHLP, the following parameter choices work well: AdLaLa: $h = 0.15, \tau_1 = \tau_2 = 10^{-4}, \gamma = 0.1, \sigma_A = 0.01, \epsilon = 0.1$. For LOL, there are fewer parameters to set; good choices appear to be $\gamma_1 = 0.01$ and $\tau_1 = 10^{-3}$ for a SHLP.   In experiments with Adam we used the hyperparameters recommended in the original article \cite{adam}, namely $\beta_1=0.9$, $\beta_2=0.999$, $\epsilon_{\rm adam} = 10^{-8}$, and did not change the learning rate throughout training.

\subsection{Comparison of classifiers} 
\label{subsec:models_classifiers}
\begin{figure}[htp]
\centering
\includegraphics[width=1.54in]{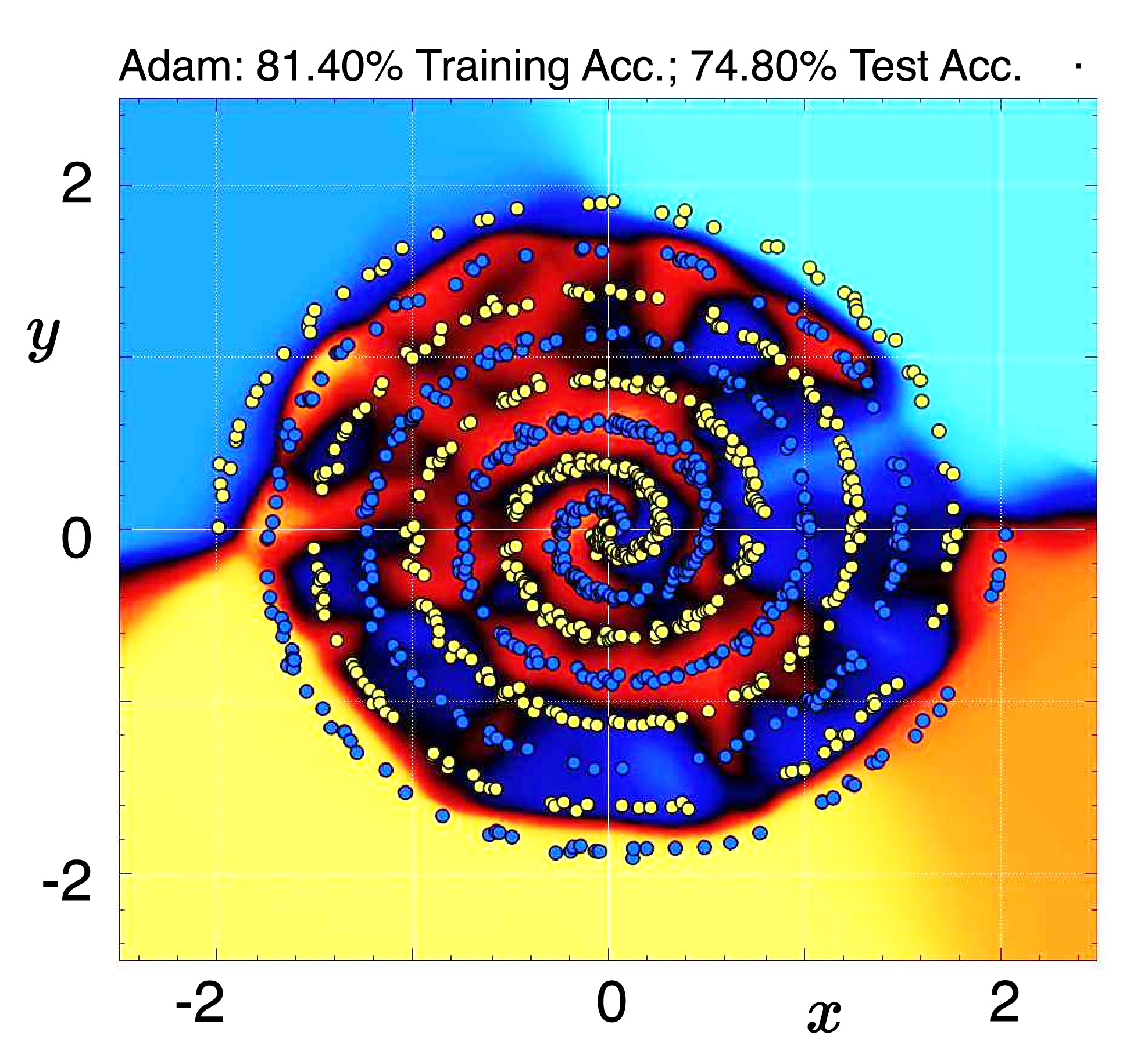}
\includegraphics[width=1.53in]{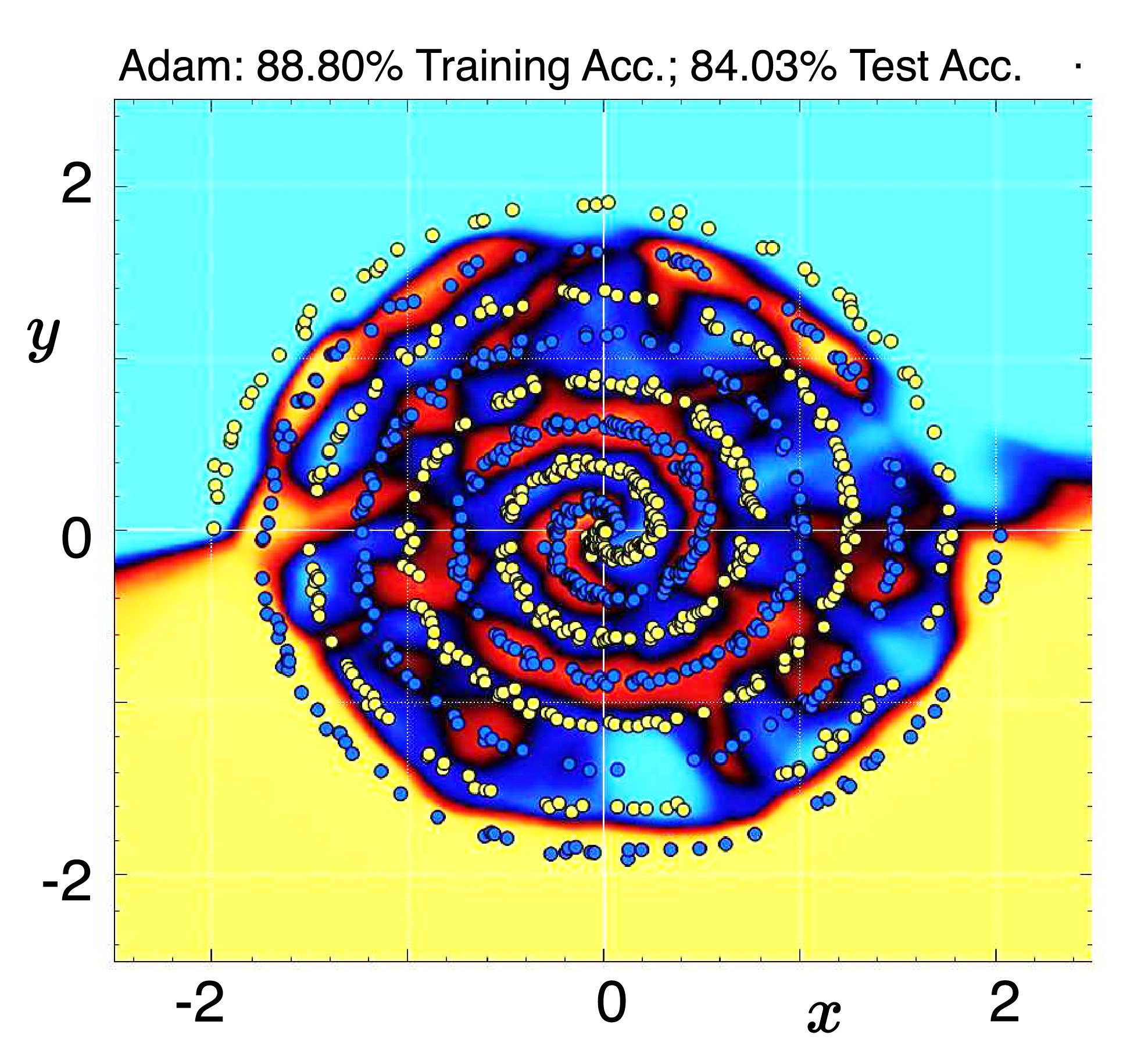}
\includegraphics[width=1.75in]{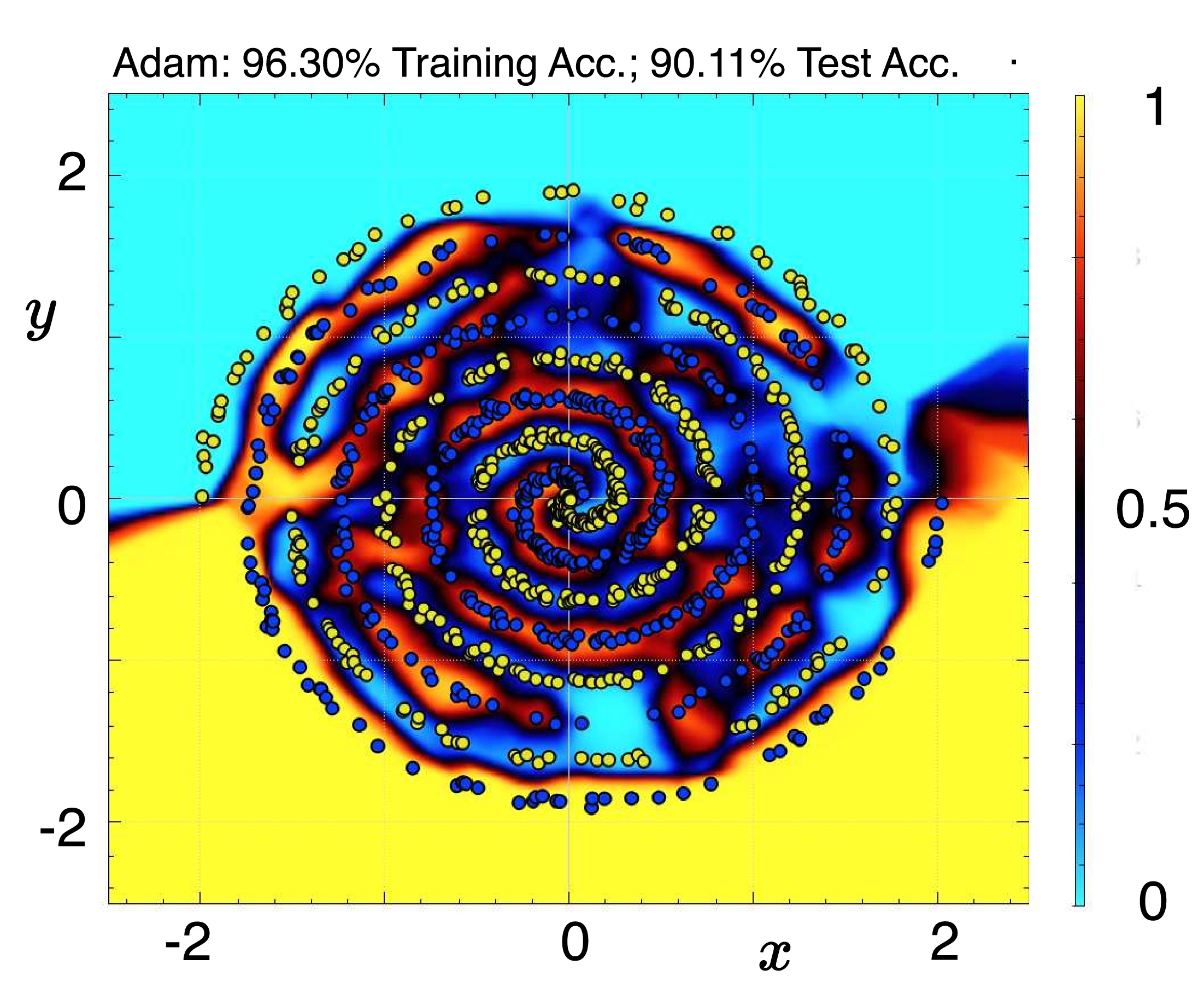}\\
\includegraphics[width=1.54in]{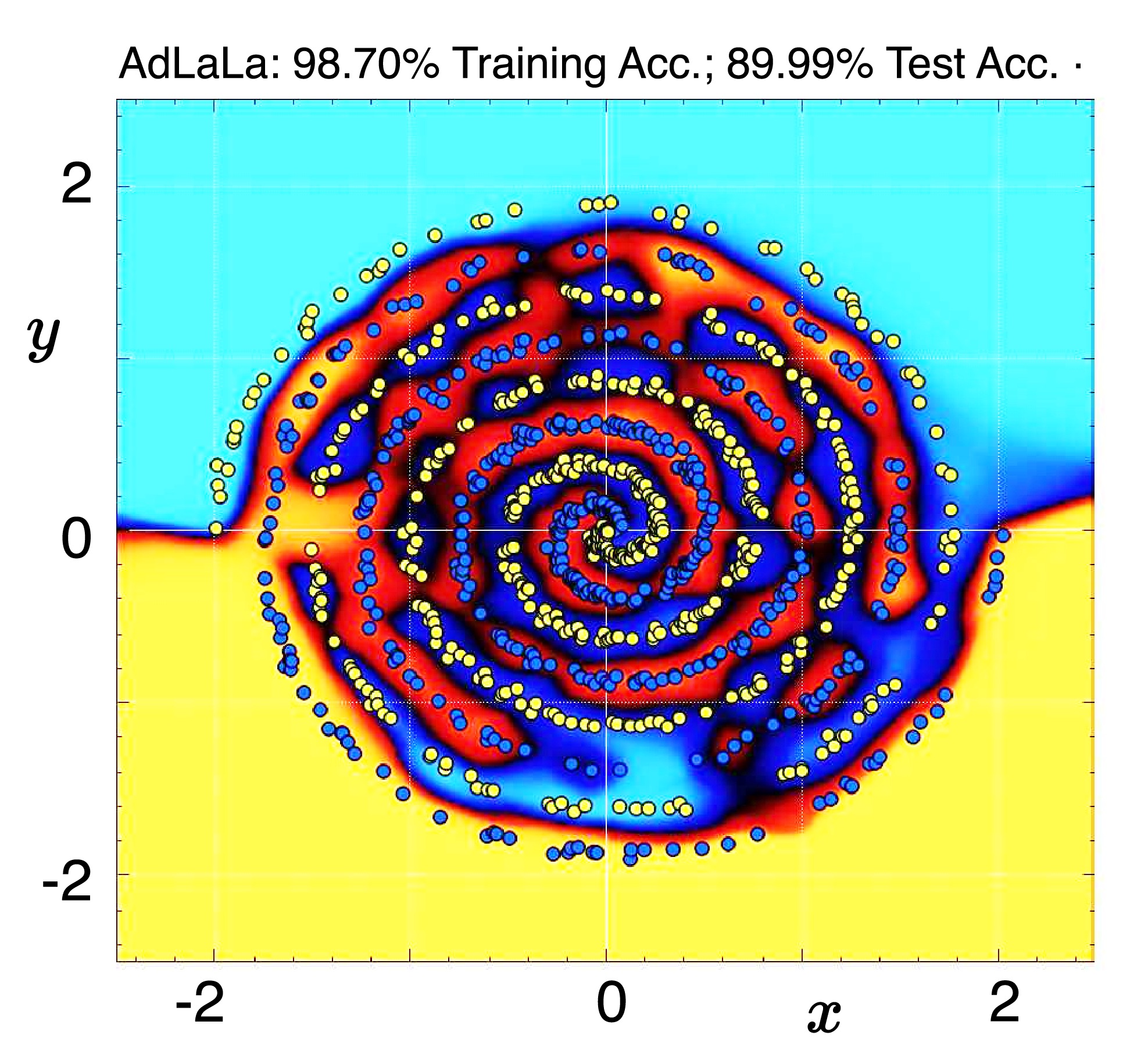}
\includegraphics[width=1.53in]{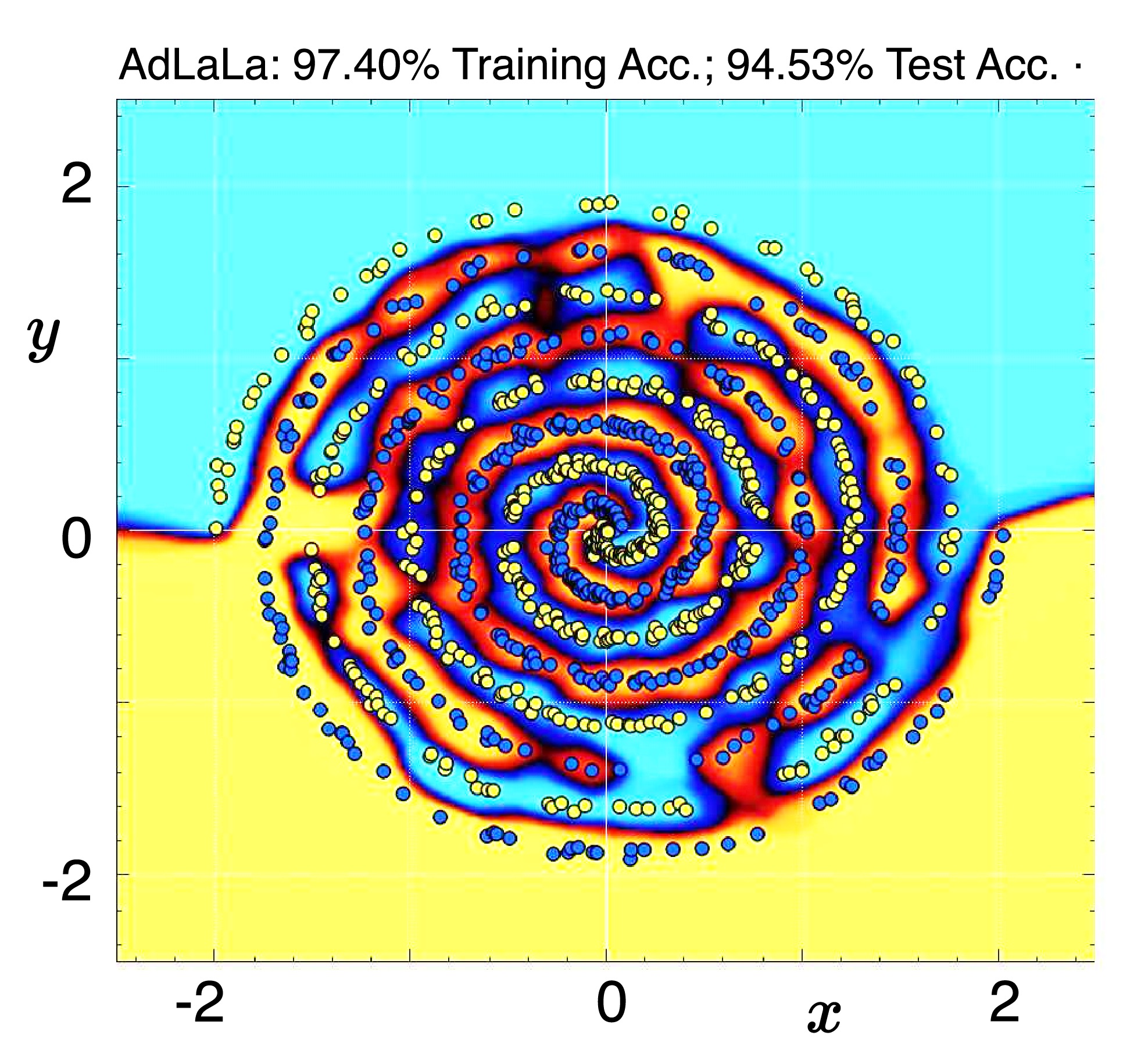}
\includegraphics[width=1.75in]{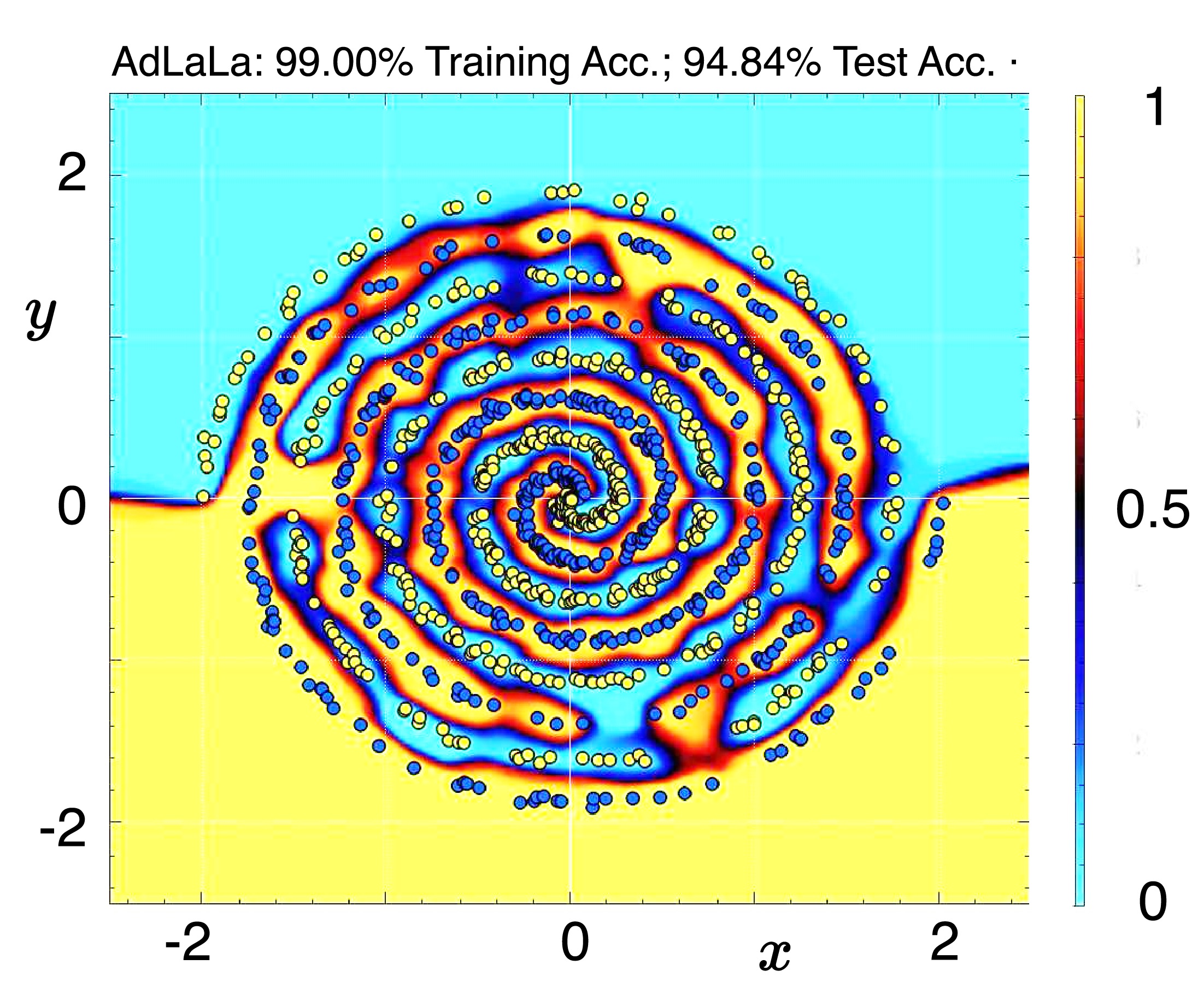}
\caption{Comparison of classifiers for a 500-node SHLP on 4-turn spiral data (with $a = 2, b = 4, c = 0.02, p = 1$ in Eq. \eqref{spiraleqn}) generated by Adam (top row) vs  AdLaLa (bottom row). For Adam the stepsize used was $h = 0.005$. Adam was initialized with Gaussian weights with standard deviation 0.5. For AdLaLa the parameters were $\epsilon = 0.1$, $\tau_1 = 0.0001$, $\sigma_A = 0.01$, $\gamma_2 = 0.03$, $\tau_2 = 0.00001, h = 0.1$. Weights were initialized as Gaussian with standard deviation 0.01.  For both methods we used 2\% subsampling per step. From left to right in each row:  20K steps (400 epochs); 40K steps (800 epochs); 60K steps (1200 epochs). For visualization the classifier was averaged over the last 10 steps of training.}\label{classifierAdamvsAdlala}
\end{figure}

The enhanced performance of AdLaLa vs Adam for the difficult 4-turns spiral dataset can be seen by comparing the classifiers they each produce (see Fig. \ref{classifierAdamvsAdlala}). We observe that the AdLaLa classifiers are far better resolved: they have lower loss and higher test accuracy, and are, moreoever, smoother.

\subsubsection{Evolution of the weights and classifier boundaries during training}
We studied the development of the classification boundary between the two spiral classes as training progresses. We observe that no matter which optimizer has been used, all of the classifiers tend to distinguish the outer part of the two spirals first, before slowly filling in the classification plane inwards. Early on in the training, there are weights which are assigned a specific role in fixing the shape of the classification boundary in the outer part of the spirals. These weights typically keep the same role throughout training.

We also isolated the effect of single data points on the training procedure. We distinguish data points from the center of the spirals and data points in the outer part of the spirals. 
We observed that for SGD data points from the outer part caused a larger change in the weight/bias values than the inner data points (at least initially, it typically changes after 2000 steps or so). For AdLaLa, however, the inner and outer data points affected the weights more or less equally from the outset. 

\subsection{Thermodynamic parameterization methods can have high accuracy and rapid convergence}\label{subsec:numerics_accuracy}
We provide evidence that our methods LOL and AdLaLa are able to converge more rapidly to a low test-loss parameterization than standard optimizers such as SGD, SGLD or Adam, for the spirals and trigonometric datasets. Our methods also perform competitively on the MNIST dataset compared to standard optimizers, but do not significantly outperform other methods for this problem. 

\begin{figure}[htp]
\centering
\includegraphics[width=2.4in]{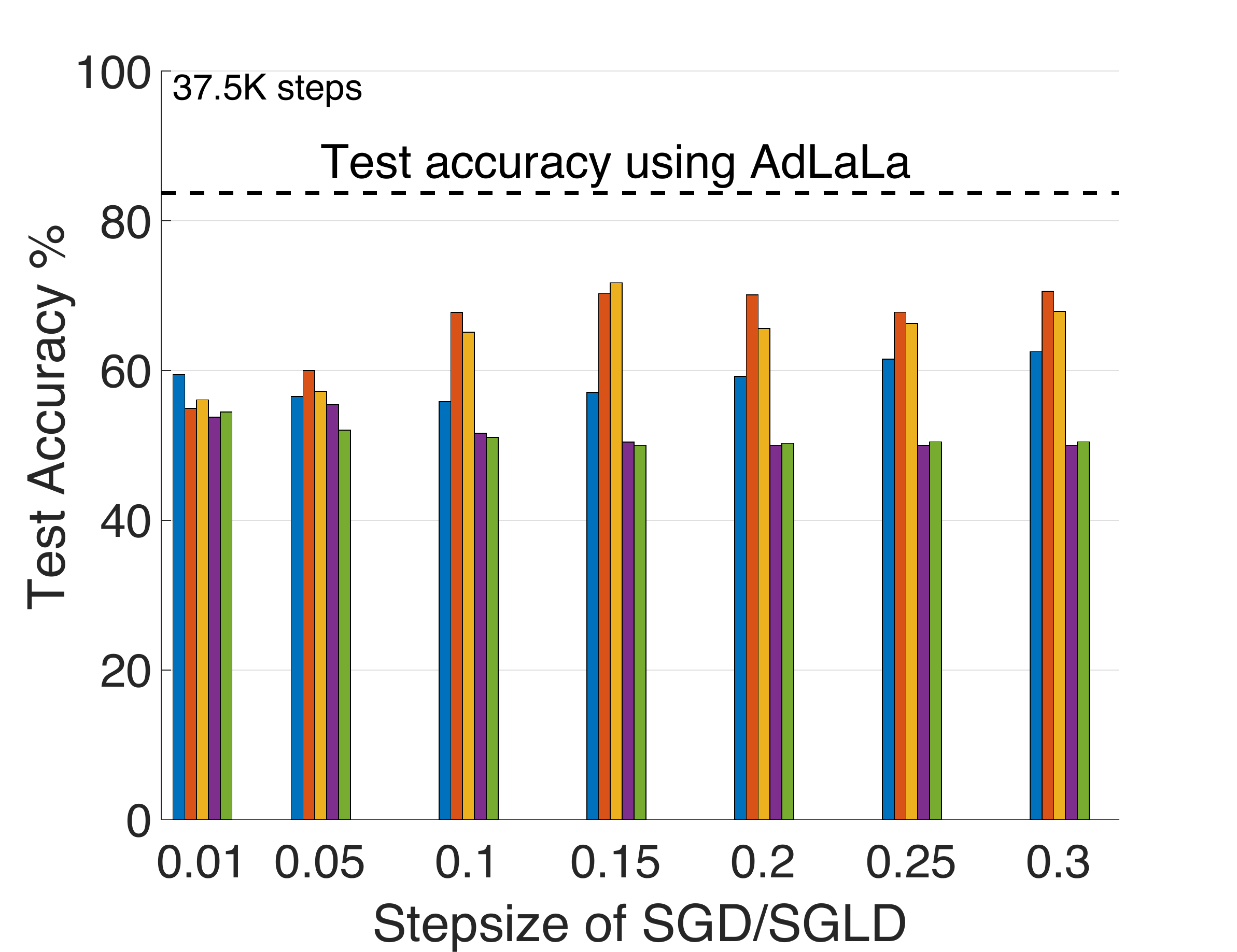}
\includegraphics[width=2.4in]{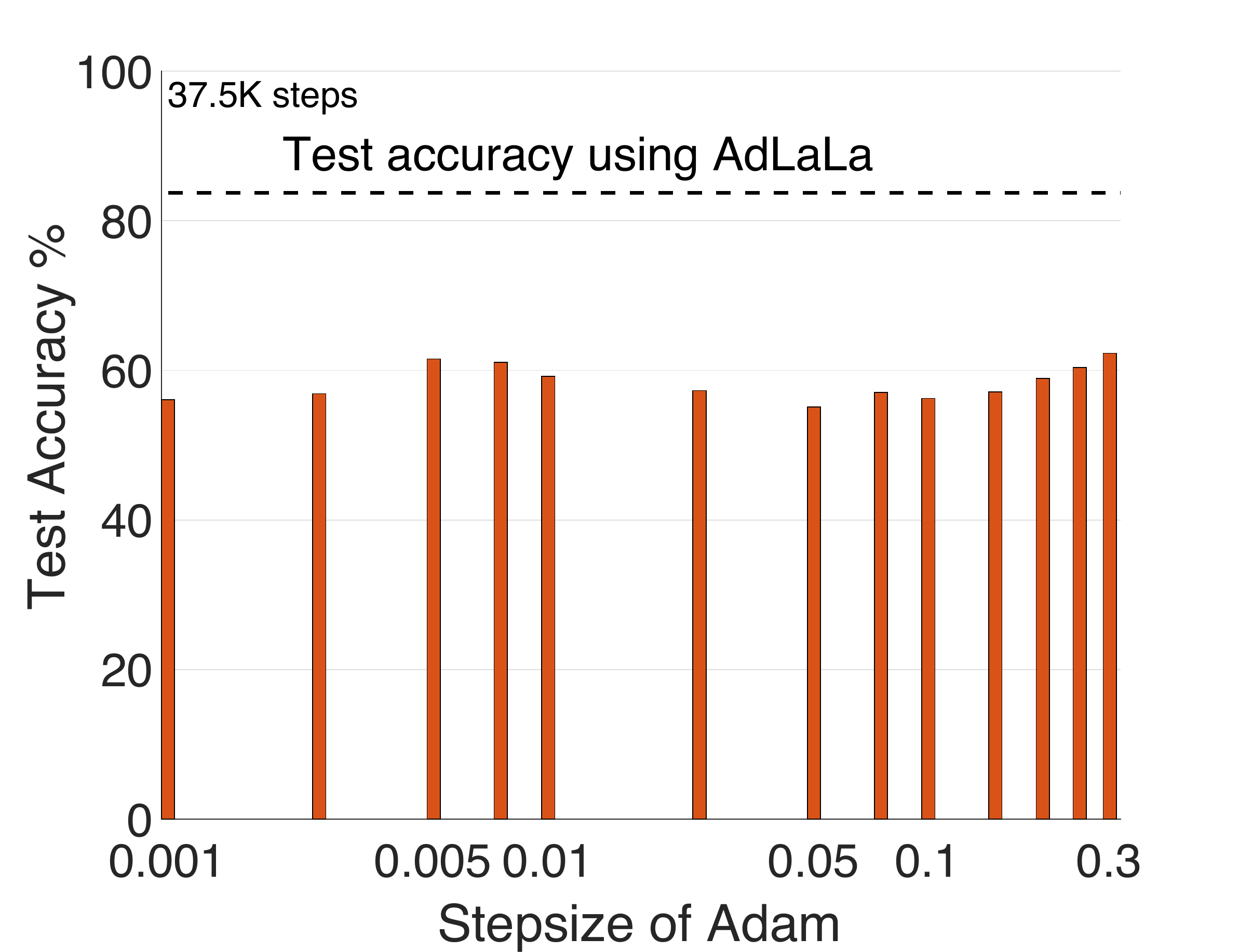}
\caption{AdLaLa (black dotted horizontal line in both figures) consistently outperforms SGD, SGLD (left figure) and Adam (right figure) for the spiral 4-turn dataset. The different bars in the left figure indicate SGLD with different values of $\sigma$, namely $\sigma = 0$ (blue, this is standard SGD), $\sigma$ = 0.005 (red), $\sigma$ = 0.01 (yellow), $\sigma$ = 0.05 (purple), $\sigma$ = 0.1 (green). Whereas the set of parameter values for AdLaLa is fixed, the parameters of the other methods were varied to show the general superiority of AdLaLa. The results were averaged over multiple runs and the same initial conditions were used for all runs. The parameters used for AdLaLa were  $h = 0.25, \tau_1 = \tau_2 = 10^{-4}, \gamma = 0.1, \sigma_A = 0.01, \epsilon = 0.05$.}
\label{fig:compare4turn}
\end{figure}

In the following experiment we show the superiority of our AdLaLa method on the spirals dataset by fixing the parameters of AdLaLa, but varying the parameters of the other methods (at this point we only varied the stepsize for Adam, not its default parameters, i.e. we did not change the decay rates for the moving averages of the first and second moments). We show that AdLaLa consistently outperforms the other methods in terms of convergence rate. The experiments were performed using a neural network with a single hidden layer consisting of 100 nodes, 1000 test data, 1000 training data and 2\% subsampling. We present comparisons for the spirals 4-turn dataset (Fig. \ref{fig:compare4turn}). We ran similar comparisons for easier 3-turn spiral data and observed similar trends. The amount of subsampling did not seem to affect the results much.

\begin{figure}[htp]
\begin{center}
    \includegraphics[width=4.5in]{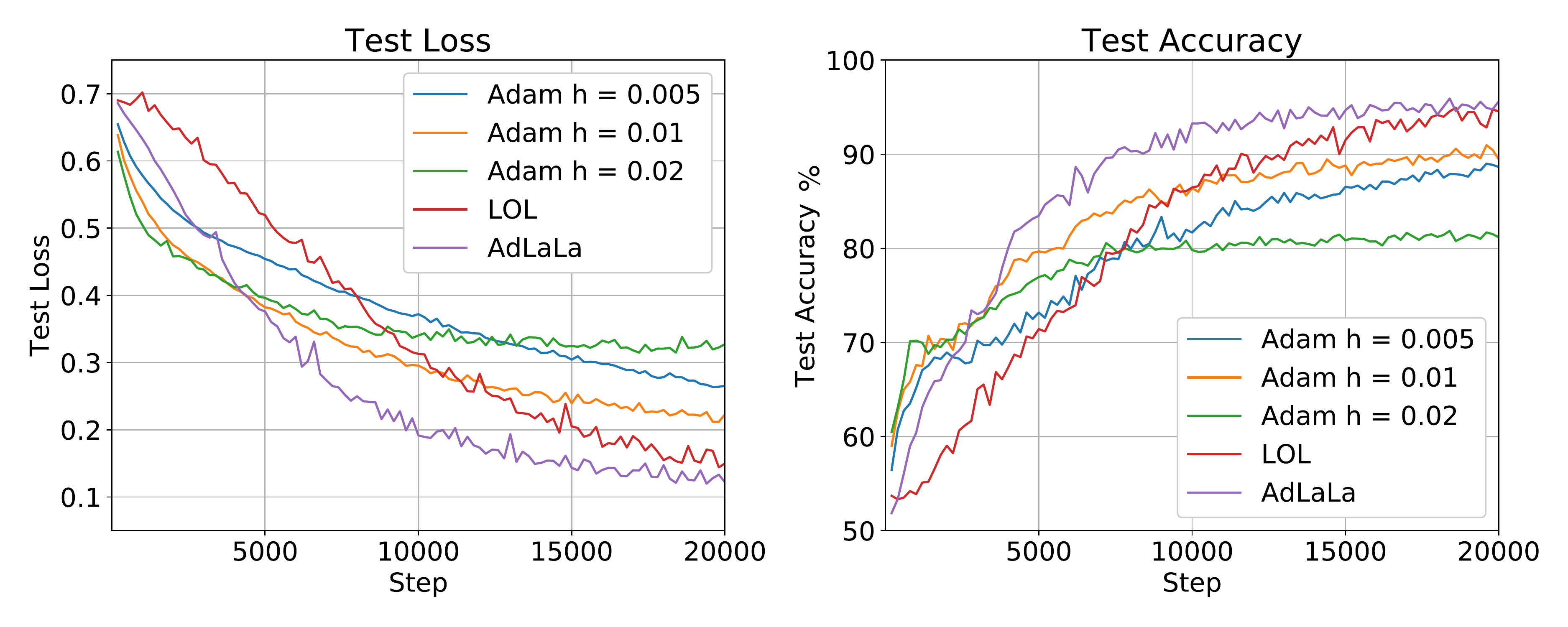}
\caption{Test loss/accuracy obtained for planar trigonometric data (with a = 6 in Eq. \eqref{trigeqn}) using different optimizers and a 100 node SHLP, 
1000 test data, 1000 training data and 5\% subsampling. The parameters for LOL are set to $h = 0.1, \gamma_1 = 0.01, \tau_1 = 10^{-3}$. For AdLaLa we used parameters: $ h = 0.2, \tau_1 = \tau_2 = 10^{-4}, \gamma = 10, \sigma_A = 0.001, \epsilon = 0.1$.}
\label{comparemethodsCosSin}
\end{center}
\end{figure}

We also show for the planar trigonometric example with $a = 6, b = 1, c = 0.02$ in Eq. \eqref{trigeqn} that our methods, LOL and AdLaLa, outperform Adam in terms of convergence rate (see Fig. \ref{comparemethodsCosSin}). Even at its (for this example) optimal time stepsize of $h = 0.01$ Adam is almost three times as slow as AdLaLa in obtaining 90\% test accuracy. 

\begin{figure}[htp]
\begin{center}
    \includegraphics[width=4.5in]{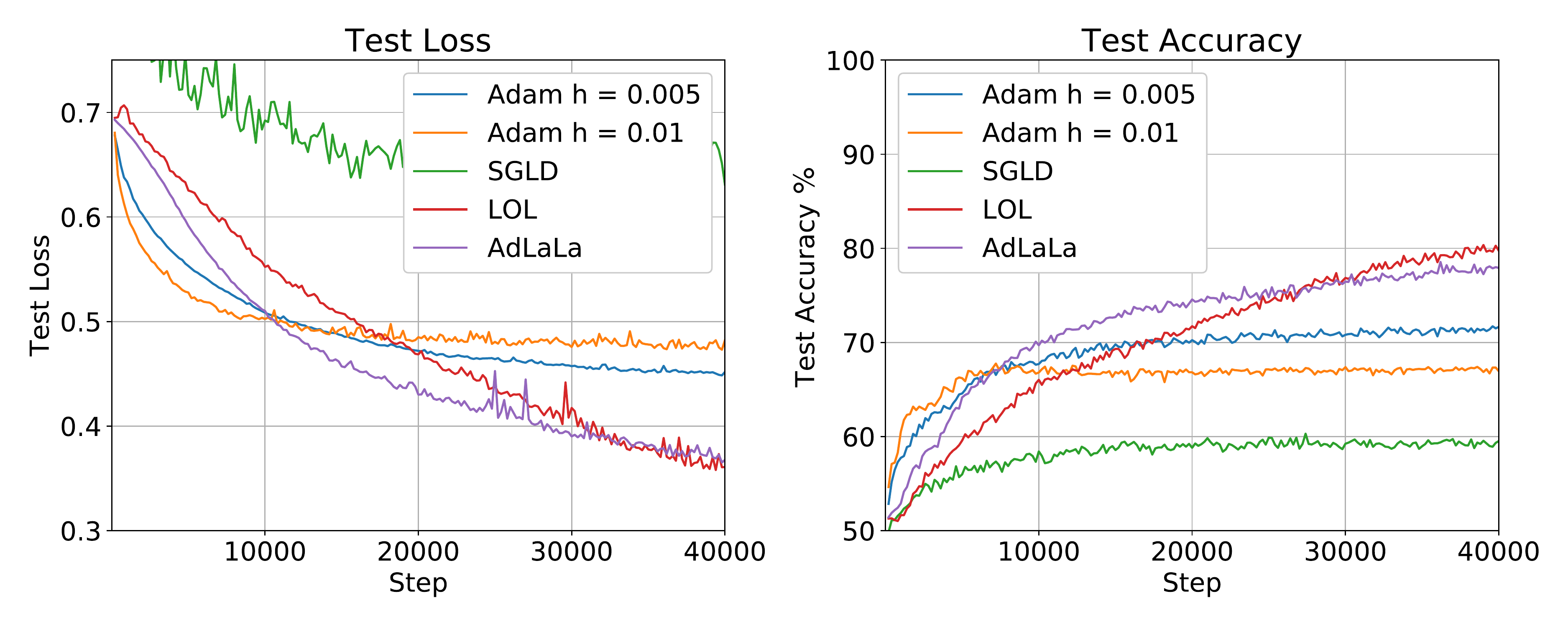}
\caption{Test loss/accuracy obtained for planar trigonometric data (with a = 10 in Eq. \eqref{trigeqn}) with a 100 node SHLP, which was parameterized using different optimizers. The results were averaged over 20 runs. Hyperparameters settings: for LOL: $h = 0.1, \gamma_1 = 0.01, \tau_1 = 10^{-3}$; for AdLaLa: $ h = 0.1, \tau_1 = \tau_2 = 10^{-4}, \gamma = 5, \sigma_A = 0.001, \epsilon = 0.1$; for SGLD: $h = 0.1, \sigma = 0.01$.}\label{comparemethodsCosSinharder}
\end{center}
\end{figure}

In our tests on a harder example, which exhibits more crossings of the two data classes, namely $a = 10$ in Eq. \eqref{trigeqn}, Adam was never able to reach the accuracy that LOL and AdLaLa obtain (see Fig. \ref{comparemethodsCosSinharder}). Its progress slows down rapidly and halts completely after 40,000 steps. After 100,000 steps its maximum test accuracy is still around 73\%. SGLD is not able to compete at all. 
\newpage

\subsection{Thermodynamic parameterization methods can reduce overfitting} \label{subsec:numerics_overfitting}
In this section we evaluate the robustness of our algorithms to overfitting. Overfitting is defined as the increase in test loss over time as the optimizer ``overfits" on the provided training data and therefore has a reduced generalization performance. To emphasize the overfitting effect, we shall decrease the amount of our training data relative to our test data, namely we shall use 200 training data points vs 4000 test data points. We also increase the noise level in our 2-turn spiral dataset to $c = 0.1$ and use a 500 node SHLP. 

\begin{figure}[htp]
\centering
\includegraphics[width=4.3in]{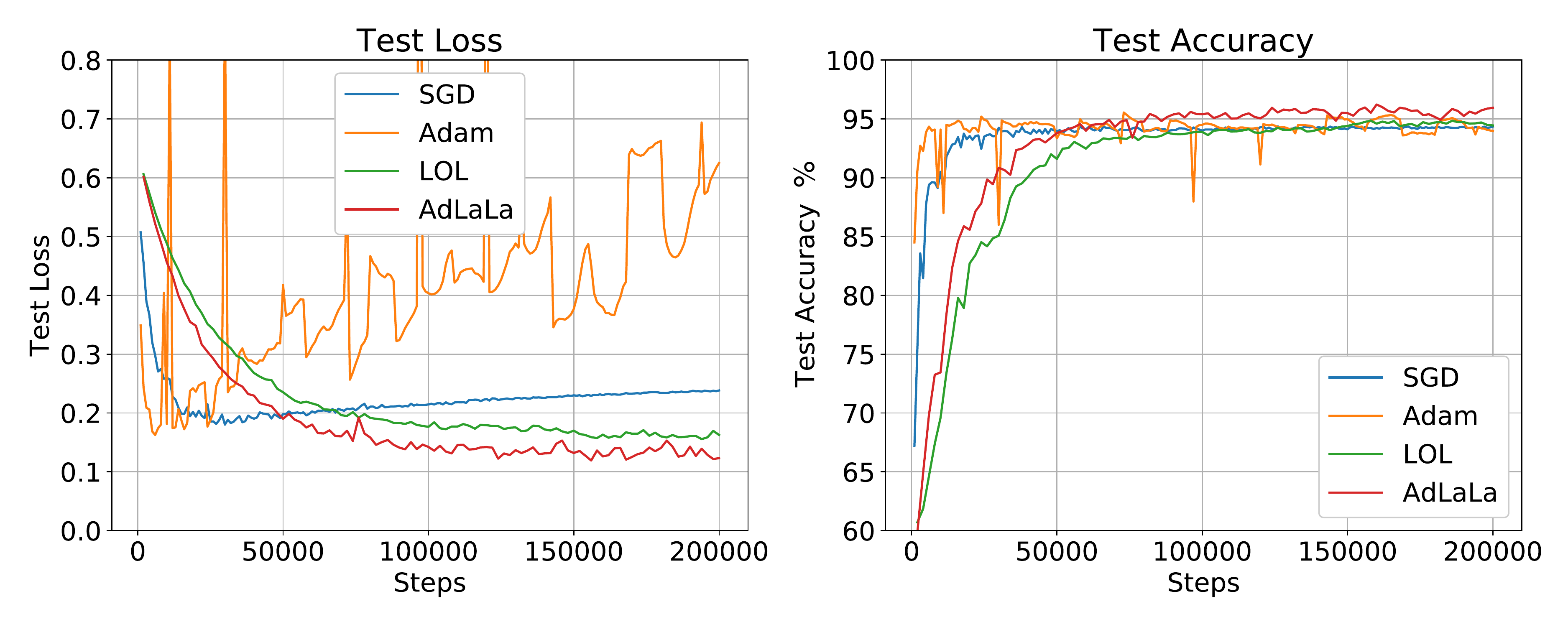}
\caption{Obtained while training a 500-node SHLP on the 2-turn spiral (with $c = 0.1$ in Eq. \eqref{spiraleqn}). We used $h_{\text{SGD}} = 0.1, h_{\text{Adam}} = 0.005$, for LOL: $h = 0.1, \gamma_1 = 1, \tau_1 = 10^{-6}$, for AdLaLa: $ h = 0.1, \tau_1 = 10^{-4}, \tau_2 = 10^{-8}, \gamma = 1000, \sigma_A = 0.01, \epsilon = 0.1$.}\label{fig:overfit}
\end{figure}

In Fig. \ref{fig:overfit} one observes that SGD clearly overfits in the sense that after a certain time its test loss monotonically increases with the number of steps. In contrast, LOL with a large enough value of $\gamma_1$ can be shown to not exhibit this behaviour. The same can be said for AdLaLa, but only after a careful selection of the method's parameter values. We note that for these parameter settings LOL and AdLaLa are slower in reaching the desired test and training accuracy, but this leads to more stability later on in the training process and limits the need for early stopping techniques. We do not claim that our methods universally counter overfitting, merely that they allow more flexibility which can lead to increased robustness to overfitting.

\subsection{Thermodynamic parameterization methods are more robust than ADAM and SGD}
\label{subsec:numerics_robustness}

\begin{figure}[htp]
\begin{center}
\includegraphics[width=4.5in]{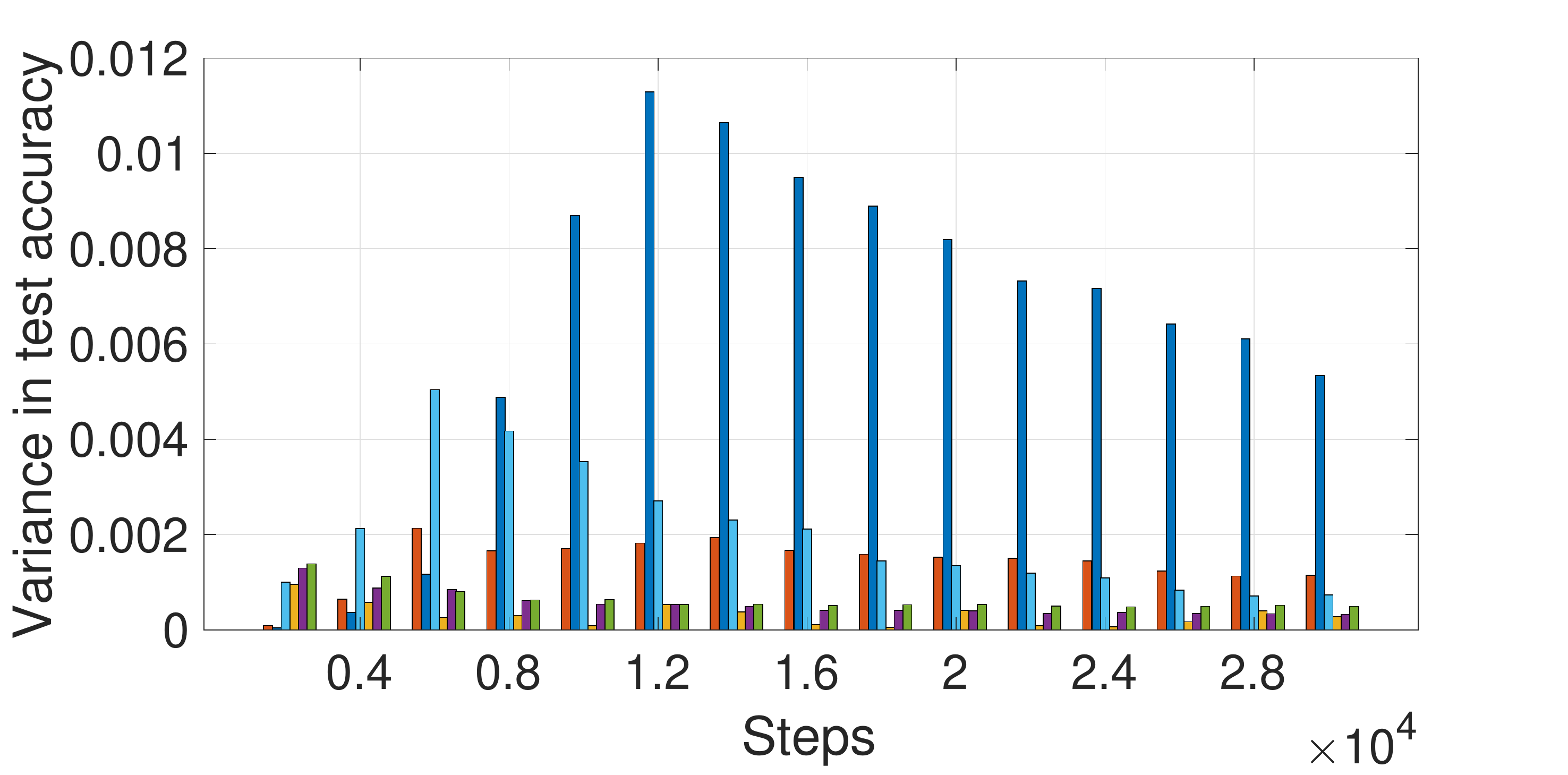}
\includegraphics[width=4.5in]{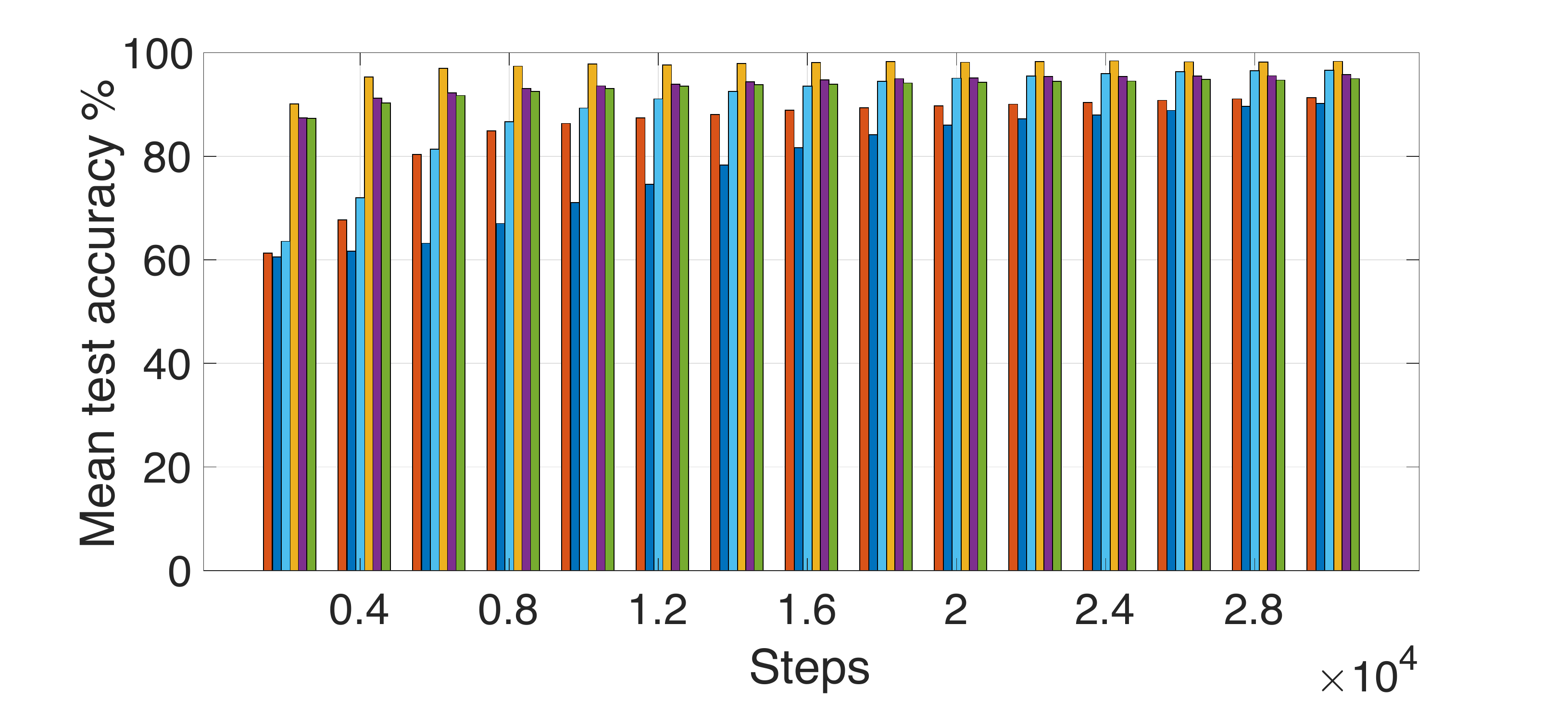}
\caption{Variance (top) and mean (bottom) in test accuracies obtained over 100 runs on the two-turn spiral problem using SGD (red) with $h = 0.25$, Adam (dark blue) with $h = 0.005$ and 0.01 $\cdot \mathcal{N}(0,1)$ initialization for the weights, Adam (light blue) with $\mathcal{U}(-1/\sqrt{N_{in}},1/\sqrt{N_{in}})$ (standard PyTorch) initialization for the weights (where $N_{in}$ is the number of inputs to the layer), LOL (yellow) with $h = 0.25, \gamma_1 = 0.01, \tau_1 = 10^{-3}$, and AdLaLa (purple) with $h = 0.25, \tau_1 = \tau_2 = 10^{-4}, \gamma = 0.5, \sigma_A = 0.01, \epsilon = 0.1$ with Gaussian initialization, AdLaLa (green) with standard PyTorch initialization. We used a 20 node SHLP, 500 training data and 2\% subsampling.}\label{fig:variance}
\end{center}
\end{figure}

We show that for the two turn spiral problem, Adam and SGD have a larger variance in their test accuracies over different runs than AdLaLa or LOL. We ran each of the optimizers 100 times and plotted the variance of the obtained test accuracies (see Fig. \ref{fig:variance}). 

The behaviour of Adam is highly dependent on the choice of initialization, while AdLaLa is less sensitive. We illustrate this by using both the standard PyTorch initialization \cite{KaimingUniform, Pytorch} for the weights (Adam is light blue and AdLaLa is green in Fig. \ref{fig:variance}) and using a Gaussian initialization (Adam is dark blue and AdLaLa is purple in Fig. \ref{fig:variance}). We also use Gaussian initialization for the other methods: SGD (red) and LOL (yellow). We observe that our methods -- yellow (LOL) and green/purple (AdLaLa) in Fig. \ref{fig:variance} -- have a much lower variance in their obtained test accuracies than Adam (with both initializations) and SGD. 

\subsection{Role of additive noise $\sigma_A$ in AdLaLa}

\begin{figure}[htb] 
\centering
    \includegraphics[width=.6\textwidth]{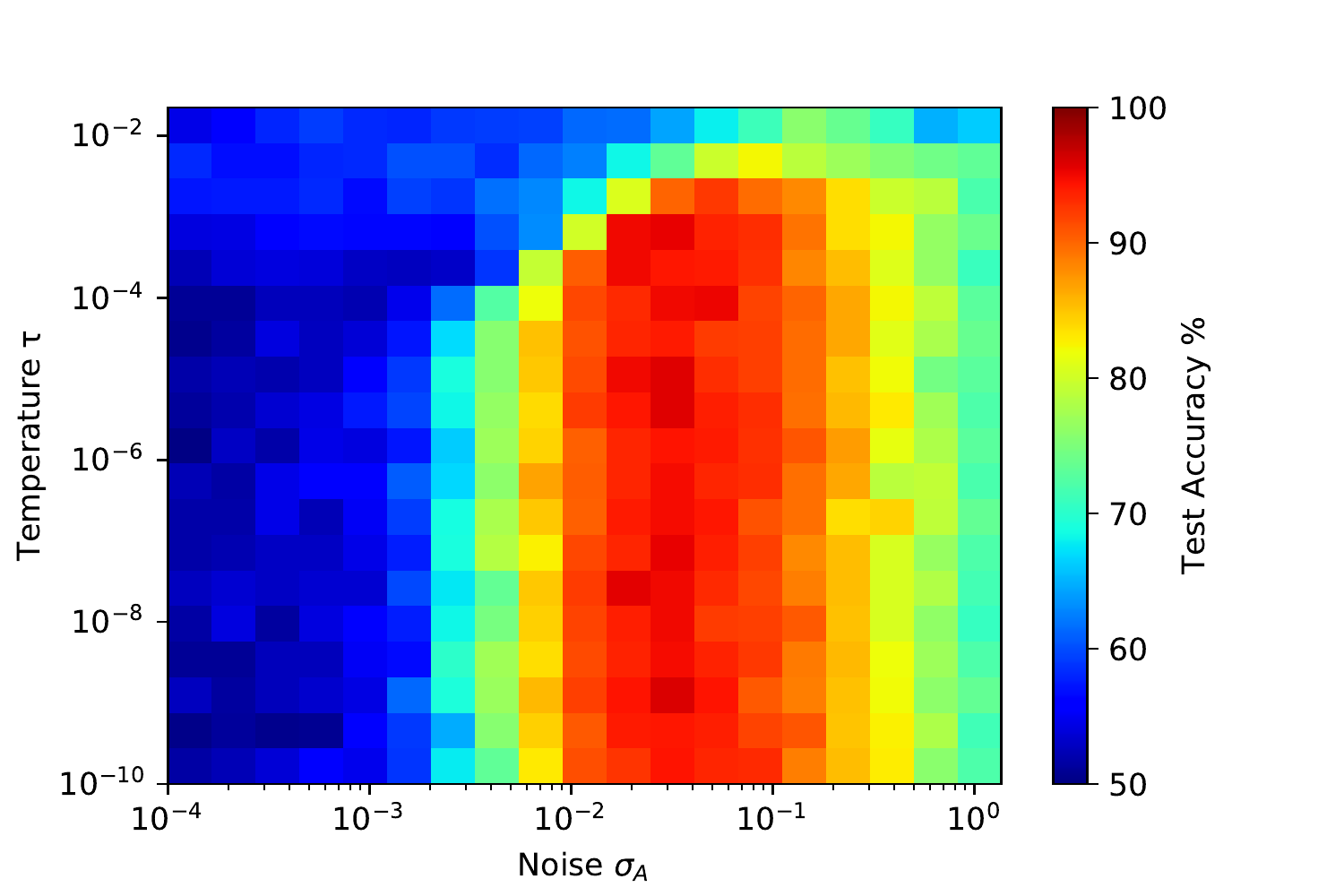} 
    \caption{We run the AdLaLa scheme on an SHLP with 100 hidden nodes on the four turn spiral problem. Pixels indicate the average test accuracy with corresponding parameters, from ten independent runs, where $\gamma_2=0.03$, $\epsilon=0.1$, $\tau_2=10^{-8}$, and $h=0.1$.} \label{fig:adlala_series}
 \end{figure}

As we can expect that gradient subsampling will introduce noise into the system, it is not immediately clear what the benefit of including additive noise is in the AdLaLa scheme (see Eq. \eqref{eq:AdLaLa_2layer}). However, we demonstrate in Fig. \ref{fig:adlala_series} that choosing an appropriate noise strength $\sigma_A>0$ can provide faster convergence to high quality minima.

We run experiments on classifying the four turn spiral problem using an SHLP with 100 hidden nodes. We draw 1000 data points as training data and use $2\%$ subsampling for computing the gradient, with the test accuracy computed from 1000 independently drawn points. The parameters in the second layer are fixed for all experiments at $\gamma_2=0.03$ and $\tau_2=10^{-8}$, with $\epsilon=0.1$. We look at the performance of the scheme for different values of $\tau_1$ and $\sigma_A$ by plotting the test accuracy (averaged over ten independent runs) after 50K steps with $h=0.1$.

The results in Fig. \ref{fig:adlala_series} demonstrate that there is a broad range (at least an order of magnitude) where using additive noise significantly improves the performance of the classifier. We observe that reducing the strength of the additive noise too much (choosing $\sigma_A<10^{-3}$ for example), or removing it entirely by setting $\sigma_A=0$, gives very poor results for the overall classification, with results no better than random noise.  By contrast, we are able to recover near $100\%$ accuracy for the same computational cost and with the same parameters by including additive noise of sufficient strength (for example choosing $\sigma_A=0.04$).

The performance of the scheme seems relatively agnostic to the choice of target temperature parameter $\tau_1$, provided it is sufficiently small. At too large a temperature the system is prevented from converging to an energy minima, leading to poor classification accuracy.

\subsection{Role of Temperature in Partitioned Schemes}

\begin{figure}[htp]
\centering
\includegraphics[width=1.5in]{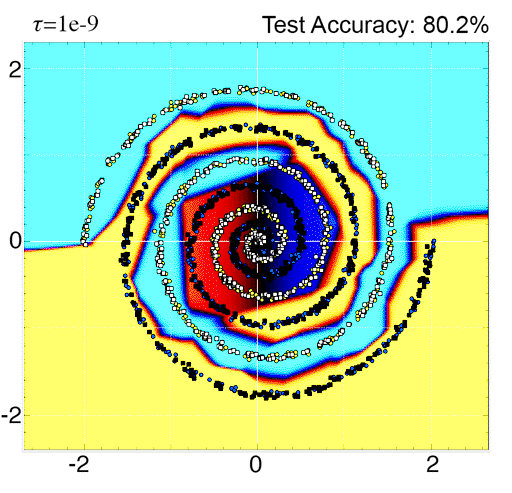}
\includegraphics[width=1.5in]{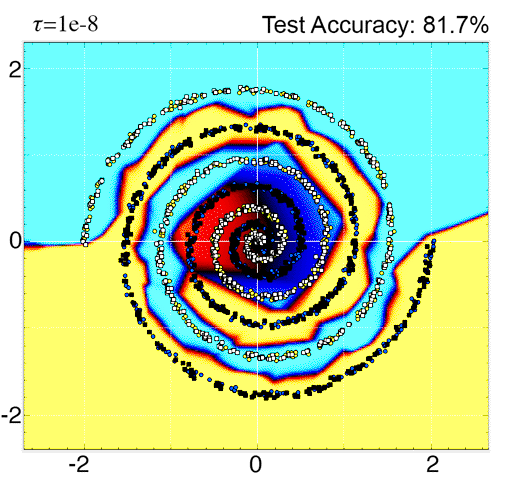}
\includegraphics[width=1.5in]{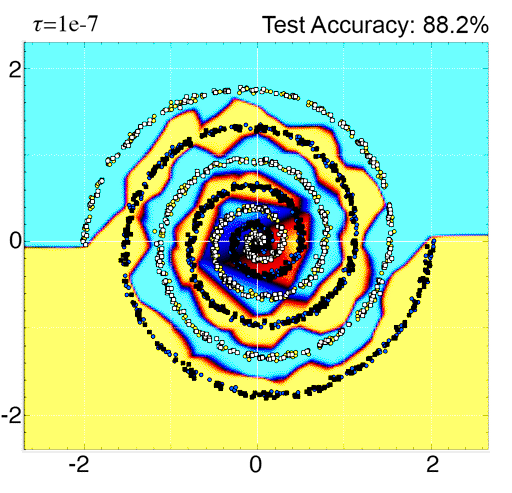}\\
\includegraphics[width=1.5in]{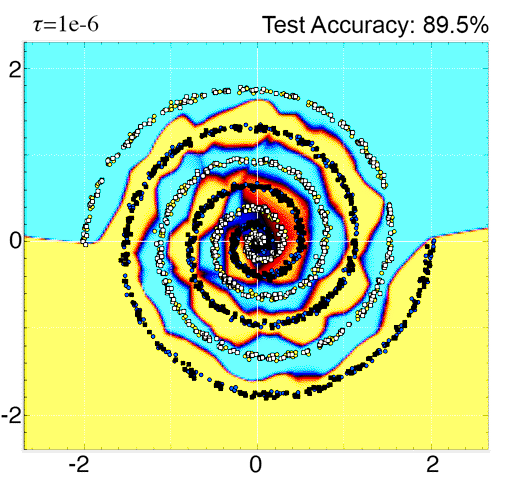}
\includegraphics[width=1.5in]{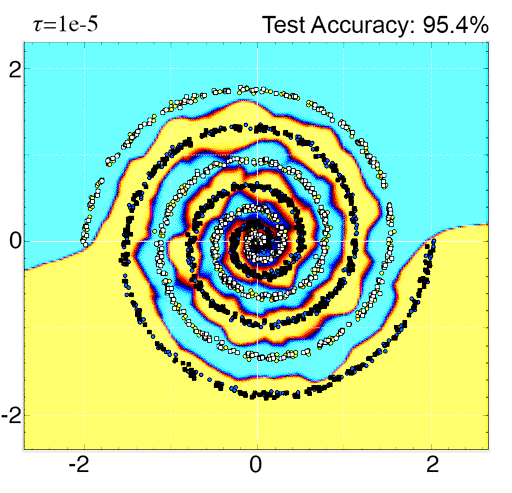}
\includegraphics[width=1.5in]{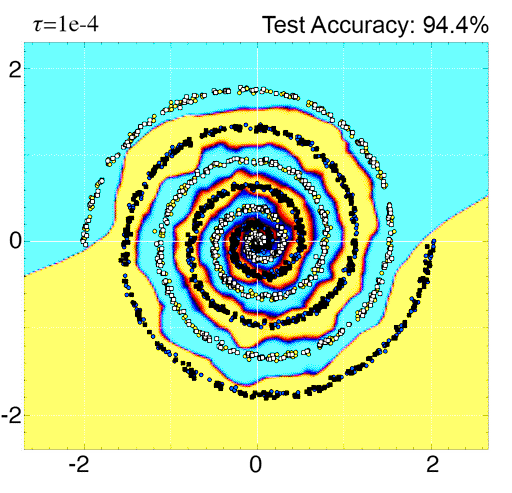}
\caption{Comparison of classifiers for a 200-node SHLP on 4-turn spiral data generated by LOL with different temperature values.   The friction was set at 1 in all experiments and 50,000 steps were performed with stepsize 0.8 (similar to large stepsizes used in SGD).  Here performance increased with increasing $\tau$ until $\tau=0.00001$ after which it began to decrease.  (The method is unusable already for $\tau =0.001$.)  }\label{fig:spiral_lol_tau}
\end{figure}

In subsection \ref{roleoftemp}, we showed in Fig. \ref{fig:baoab_trig} 
that the Langevin schemes could be more accurate when used with higher temperature.  We close this series of numerical experiments with a demonstration using the 4-turn spiral data that the LOL method similarly is more accurate at modest temperatures (i.e. there is a band of temperature for which LOL performance improves), see Fig. \ref{fig:spiral_lol_tau}.

\section{Conclusion and Outlook}
\label{sec:conclusion}

We have presented a new approach to parameterization of neural networks which can, in challenging data classification problems, accelerate convergence and provide improved test accuracy.  The use of additive noise to supplement gradient noise was already proposed in previous works of other authors.  We draw on this, by combining it with state-of-the-art principles for sampling algorithms coming from molecular dynamics and deploy partitioned algorithms that substantially improve on SGD and other optimization procedures.  These new methods have other advantages -- for one thing they appear not to require additional regularization to obtain good performance (we did not use regularization in our experiments). Another advantage
is that the stochastic methods, namely partitioned Langevin, LOL and AdLaLa, do not require complex initialization in the cases we studied. In fact, we initialized them frequently from zero initial weights and momenta and sometimes using built in training package procedures such as that in DLIB and PyTorch. This did not seem to significantly impact their performance.

The implementation of many of these schemes is straightforward, although obviously any major code project will require substantial investment of time and planning if the result is to be reliable software which is  scalable to large data sets and network sizes.    As preliminary groundwork, we have already released a software package called TATi (Thermodynamic Analytics Toolkit) which implements Langevin dynamics methods on the TensorFlow platform.\footnote{TATI is available within the Python Package installer and can be installed in a few minutes using the command {\tt pip install tati}.}   We hope to extend this software package in the near future to also implement the more complex partitioned methods discussed in the article.  

In terms of future directions for research, we mention several important challenges. First, the experiments of this article have all focussed on a limited collection of toy data sets, specifically classification problems for planar data.  These present some difficulty for common training methods, so they are a good first step, but it is natural to look next at some state-of-the-art challenges such as arise in large scale image classification or natural language processing.   Second, the power of these methods will not be fully recognized by the field until the results are demonstrated in deep networks, which are increasingly popular for machine learning applications due to better accuracy and generalization capabilities.  We have in fact implemented the methods already for such networks and we expect to publish a paper on this topic soon.    

Finally, we highlight the improved generalization properties of the models trained using our methods, as demonstrated in our experiments.     Nowhere is the problem of poor generalization more acute than in the study of streaming data, where the continual perturbation of the data leads to aging of parameter sets which can necessitate frequent costly reparameterization. 
We therefore look to this topic for a rich source of problems to test out our methods in the future.

\section*{Acknowledgements}
The authors wish to thank John Chodera, Jason Frank, Anton Martinsson, Klaus-Robert M\"uller, Gabriel Stoltz, Amos Storkey, and Jonathan Weare for helpful discussions during the preparation of this manuscript.  The work of Benedict Leimkuhler and Charles Matthews was supported by the Engineering and Physical Sciences Research Council (EPSRC) under EP/P006175/1.  Benedict Leimkuhler is also a fellow of the Alan Turing Institute which is funded by grant EPSRC EP/N510129/1 and has benefited from this fellowship in the development of this work.
Tiffany Vlaar is  supported by The Maxwell Institute Graduate School in Analysis and its Applications, a Centre for Doctoral Training funded by the UK Engineering and Physical Sciences Research Council (grant EP/L016508/01), the Scottish Funding Council, Heriot-Watt University and the University of Edinburgh. This work has made use of the resources provided by the Edinburgh Compute and Data Facility  (ECDF)\footnote{See http://www.ecdf.ed.ac.uk/}.

\bibliography{lit.bib}

\begin{thebibliography}{54}
\providecommand{\natexlab}[1]{#1}
\providecommand{\url}[1]{\texttt{#1}}
\expandafter\ifx\csname urlstyle\endcsname\relax
  \providecommand{\doi}[1]{doi: #1}\else
  \providecommand{\doi}{doi: \begingroup \urlstyle{rm}\Url}\fi

\bibitem[Avati et~al.(2018)Avati, Jung, S, Downing, Ng, and Shah]{med}
A.~Avati, K.~Jung, S.~Harman S, L.~Downing, A.~Ng, and N.~Shah.
\newblock Improving palliative care with deep learning.
\newblock \emph{BMC Medical Information and Decision Making}, 18\penalty0
  (122), 2018.

\bibitem[Ballard et~al.(2017)Ballard, Das, Martiniani, Mehta, Sagun, Stevenson,
  and Wales]{Ballard}
A.J. Ballard, R.~Das, S.~Martiniani, D.~Mehta, L.~Sagun, J.D. Stevenson, and
  D.J. Wales.
\newblock Energy landscapes for machine learning.
\newblock \emph{Phys. Chem. Chem. Phys.}, 19:\penalty0 12585--12603, 2017.

\bibitem[Brosse et~al.(2018)Brosse, Durmus, and Moulines]{Brosse}
N.~Brosse, A.~Durmus, and E.~Moulines.
\newblock The promises and pitfalls of stochastic gradient langevin dynamics.
\newblock \emph{NIPS}, pages 8268--8278, 2018.

\bibitem[Choromanska et~al.(2015)Choromanska, Henaff, Mathieu, Arous, and
  LeCun]{equivalentminima}
A.~Choromanska, M.~Henaff, M.~Mathieu, G.~Arous, and Y.~LeCun.
\newblock The loss surfaces of multilayer networks.
\newblock \emph{Journal of Machine Learning Research}, 38:\penalty0 192--204,
  2015.

\bibitem[Dauphin et~al.(2014)Dauphin, Pascanu, cehre, Cho, Ganguli, and
  Bengio]{saddlepoints}
Y.~Dauphin, R.~Pascanu, C.~G\"ul\c cehre, K.~Cho, S.~Ganguli, and Y.~Bengio.
\newblock Identifying and attacking the saddle point problem in
  high-dimensional non-convex optimization.
\newblock \emph{NIPS}, 2014.

\bibitem[Ding et~al.(2014)Ding, Fang, Babbush, Chen, Skeel, and Neven]{Di2014}
N.~Ding, Y.~Fang, R.~Babbush, C.~Chen, R.D. Skeel, and H.~Neven.
\newblock {B}ayesian sampling using stochastic gradient thermostats.
\newblock In \emph{Advances in neural information processing systems}, pages
  3203--3211, 2014.

\bibitem[Dolbeault et~al.(2009)Dolbeault, Mouhot, and Schmeiser]{DoMoSc09}
J.~Dolbeault, C.~Mouhot, and C.~Schmeiser.
\newblock Hypocoercivity for kinetic equations with linear relaxation terms.
\newblock \emph{C. R. Math. Acad. Sci. Paris}, 347\penalty0 (9-10):\penalty0
  511--516, 2009.

\bibitem[Duchi et~al.(2011)Duchi, Hazan, and Singer]{adagrad}
J.~Duchi, E.~Hazan, and Y.~Singer.
\newblock Adaptive subgradient methods for online learning and stochastic
  optimization.
\newblock \emph{Journal of Machine Learning Research}, 12:\penalty0 2121--2159,
  2011.

\bibitem[Durmus and Moulines(2017)]{Moulines}
A.~Durmus and E.~Moulines.
\newblock Non-asymptotic convergence analysis for the unadjusted {L}angevin
  algorithm.
\newblock \emph{The Annals of Applied Probability}, 27:\penalty0 1551--1587,
  2017.

\bibitem[ebski et~al.(2017)ebski, Kenton, Arpit, Ballas, Fischer, Bengio, and
  Storkey]{JastrzebskiStorkey}
S.~Jastrz\c ebski, Z.~Kenton, D.~Arpit, N.~Ballas, A.~Fischer, Y.~Bengio, and
  A.J. Storkey.
\newblock Three factors influencing minima in sgd.
\newblock \emph{CoRR, arXiv:1711.04623}, 2017.

\bibitem[Gardiner(2004)]{gardiner}
C.~Gardiner.
\newblock \emph{Handbook of Stochastic Methods for Physics, Chemistry, and the
  Natural Sciences}.
\newblock 3rd edn. Springer, New York, 2004.

\bibitem[Geyer(1991)]{ST}
C.J. Geyer.
\newblock Markov {C}hain {M}onte {C}arlo maximum likelihood.
\newblock \emph{Computer Science and Statistics}, 1991.

\bibitem[Glorot et~al.(2011)Glorot, Bordes, and Bengio]{ReLU2}
X.~Glorot, A.~Bordes, and Y.~Bengio.
\newblock Deep sparse rectifier networks.
\newblock \emph{AISTATS}, 2011.

\bibitem[Goodfellow et~al.(2015)Goodfellow, Vinyals, and
  Saxe]{goodfellowvinyals}
I.J. Goodfellow, O.~Vinyals, and A.M. Saxe.
\newblock Qualitatively characterizing neural network optimization problems.
\newblock \emph{ICLR}, 2015.

\bibitem[He et~al.(2015)He, Zhang, Ren, and Sun]{KaimingUniform}
K.~He, X.~Zhang, S.~Ren, and J.~Sun.
\newblock Delving deep into rectifiers: Surpassing human-level performance on
  {I}magenet classification.
\newblock \emph{Proceedings of the IEEE international conference on computer
  vision}, pages 1026--1034, 2015.

\bibitem[Herzog(2018)]{He2018}
D.P. Herzog.
\newblock Exponential relaxation of the {N}os\'e-{H}oover equation under
  {B}rownian heating.
\newblock \emph{Communications in Mathematical Sciences}, 16\penalty0
  (8):\penalty0 2231--2260, 2018.

\bibitem[Hoerl and Kennard(1970)]{ridge}
A.~Hoerl and R.~Kennard.
\newblock Ridge regression: {B}iased estimation for nonorthogonal problems.
\newblock \emph{Technometrics}, 12:\penalty0 55--67, 1970.

\bibitem[Hoover(1985)]{Ho1985}
W.~Hoover.
\newblock Canonical dynamics: Equilibrium phase-space distributions.
\newblock \emph{Phys. Rev. A.}, 31\penalty0 (3):\penalty0 1695--1697, 1985.

\bibitem[Huang et~al.(2019)Huang, Emam, Goldblum, Fowl, Terry, Huang, and
  Goldstein]{ringexample}
W.R. Huang, Z.~Emam, M.~Goldblum, L.~Fowl, J.K. Terry, F.~Huang, and
  T.~Goldstein.
\newblock Understanding generalization through visualizations.
\newblock \emph{arXiv: 1906.03291, preprint}, 2019.

\bibitem[Im et~al.(2016)Im, Tao, and Branson]{Im}
D.J. Im, M.~Tao, and K.~Branson.
\newblock An empirical analysis of deep network loss surfaces.
\newblock \emph{CoRR}, 2016.

\bibitem[Jarrett et~al.(2009)Jarrett, Kavukcuoglu, Ranzato, and LeCun]{ReLU1}
K.~Jarrett, K.~Kavukcuoglu, M.~Ranzato, and Y.~LeCun.
\newblock What is the best multi-stage architecture for object recognition?
\newblock \emph{ICCV}, 2009.

\bibitem[Jones and Leimkuhler(2011)]{JoLe2011}
A.~Jones and B.~Leimkuhler.
\newblock Adaptive stochastic methods for sampling driven molecular systems.
\newblock \emph{The Journal of Chemical Physics}, 135\penalty0 (8):\penalty0
  084125, 2011.

\bibitem[King(2009)]{dlib09}
Davis~E. King.
\newblock Dlib-ml: A machine learning toolkit.
\newblock \emph{Journal of Machine Learning Research}, 10:\penalty0 1755--1758,
  2009.

\bibitem[Kingma and Ba(2015)]{adam}
D.P. Kingma and J.~Ba.
\newblock Adam: A method for stochastic optimization.
\newblock \emph{ICLR}, 2015.

\bibitem[Kirkpatrick et~al.(1983)Kirkpatrick, Gelatt, and Vecchi]{annealing}
S.~Kirkpatrick, C.D. Gelatt, and M.P. Vecchi.
\newblock Optimization by simulated annealing.
\newblock \emph{Science}, 220:\penalty0 671--680, 1983.

\bibitem[Kushner and Yin(2003)]{Ku2003}
H.~Kushner and G.G. Yin.
\newblock \emph{Stochastic approximation and recursive algorithms and
  applications}, volume~35.
\newblock Springer Science \& Business Media, 2003.

\bibitem[Lan et~al.(2019)Lan, Liu, Zhou, and Yosinski]{LCA}
J.~Lan, R.~Liu, H.~Zhou, and J.~Yosinski.
\newblock Lca: Loss change allocation for neural network training.
\newblock \emph{arXiv: 1909.01440, preprint}, 2019.

\bibitem[Leimkuhler and Matthews(2015)]{LeMa2015}
B.~Leimkuhler and C.~Matthews.
\newblock \emph{Molecular Dynamics: With Deterministic and Stochastic Numerical
  Methods}.
\newblock Interdisciplinary Applied Mathematics. Springer, 2015.

\bibitem[Leimkuhler and Shang(2016)]{LeSh2016}
B.~Leimkuhler and X.~Shang.
\newblock Adaptive thermostats for noisy gradient systems.
\newblock \emph{SIAM Journal on Scientific Computing}, 38\penalty0
  (2):\penalty0 A712--A736, 2016.

\bibitem[Leimkuhler et~al.(2015)Leimkuhler, Matthews, and Stoltz]{LeMaSt2015}
B.~Leimkuhler, C.~Matthews, and G.~Stoltz.
\newblock The computation of averages from equilibrium and nonequilibrium
  {L}angevin molecular dynamics.
\newblock \emph{IMA Journal of Numerical Analysis}, 36\penalty0 (1):\penalty0
  13--79, 2015.

\bibitem[Leimkuhler et~al.(2019)Leimkuhler, Sachs, and Stoltz]{StSaLe2019}
B.~Leimkuhler, M.~Sachs, and G.~Stoltz.
\newblock Hypocoercivity properties of adaptive langevin dynamics.
\newblock \emph{arXiv:1908.09363, preprint}, 2019.

\bibitem[Marinari and Parisi(1992)]{ST2}
E.~Marinari and G.~Parisi.
\newblock Simulated tempering: a new {M}onte {C}arlo scheme.
\newblock \emph{Europhysics Letters}, 1992.

\bibitem[Mattingly et~al.(2002)Mattingly, A.M.Stuart, and Higham]{MaStHi2002}
J.C. Mattingly, A.M.Stuart, and D.J. Higham.
\newblock Ergodicity for {SDEs} and approximations: locally {L}ipschitz vector
  fields and degenerate noise.
\newblock \emph{Stochastic Processes and their Applications}, 101\penalty0
  (2):\penalty0 185--232, 2002.

\bibitem[Meyn and Tweedie(1993)]{MeTw1993}
S.P. Meyn and R.L. Tweedie.
\newblock Stability of {M}arkovian processes {II}: Continuous-time processes
  and sampled chains.
\newblock \emph{Advances in Applied Probability}, 25\penalty0 (3):\penalty0
  487--517, 1993.

\bibitem[Murphy(2012)]{Murphy}
K.P. Murphy.
\newblock \emph{Machine learning: A probabilistic perspective}.
\newblock MIT Press, 2012.

\bibitem[Neal(2012)]{Ne1995}
R.M. Neal.
\newblock \emph{Bayesian learning for neural networks}, volume 118.
\newblock Springer Science \& Business Media, 2012.

\bibitem[Neyshabur et~al.(2015)Neyshabur, Tomioka, and Srebro]{neyshabur2015}
B.~Neyshabur, R.~Tomioka, and N.~Srebro.
\newblock In search of the real inductive bias: On the role of implicit
  regularization in deep learning.
\newblock \emph{Proceeding of the International Conference on Learning
  Representations workshop track}, 2015.

\bibitem[Nos{\'e}(1984)]{No1984}
S.~Nos{\'e}.
\newblock A unified formulation of the constant temperature molecular dynamics
  methods.
\newblock \emph{The Journal of Chemical Physics}, 81\penalty0 (1):\penalty0
  511--519, 1984.

\bibitem[Paszke et~al.(2017)Paszke, Gross, Chintala, Chanan, Yang, DeVito, Lin,
  Desmaison, Antiga, and Lerer]{Pytorch}
A.~Paszke, S.~Gross, S.~Chintala, G.~Chanan, E.~Yang, Z.~DeVito, Z.~Lin,
  A.~Desmaison, L.~Antiga, and A.~Lerer.
\newblock Automatic differentiation in {P}y{T}orch.
\newblock 2017.

\bibitem[Pollak et~al.(2008)Pollak, Auerbach, and Talkner]{Po2008}
E.~Pollak, A.~Auerbach, and P.~Talkner.
\newblock Observations on rate theory for rugged energy landscapes.
\newblock \emph{Biophysical Journal}, 95:\penalty0 4258--4265, 2008.
\newblock URL \url{https://doi.org/10.1529/biophysj.108.136358}.

\bibitem[Roberts and Tweedie(1996)]{Roberts1996}
G.O. Roberts and R.L. Tweedie.
\newblock Exponential convergence of {L}angevin distributions and their
  discrete approximations.
\newblock \emph{Bernoulli}, 2\penalty0 (4):\penalty0 341--363, 1996.

\bibitem[Sachs et~al.(2017)Sachs, Leimkuhler, and Danos]{SaLeDa2017}
M.~Sachs, B.~Leimkuhler, and V.~Danos.
\newblock Langevin dynamics with variable coefficients and nonconservative
  forces: from stationary states to numerical methods.
\newblock \emph{Entropy}, 19:\penalty0 647, 2017.

\bibitem[Sch{\"u}tt et~al.(2017)Sch{\"u}tt, Arbabzadah, S.~Chmiela, and
  Tkatchenko]{Sc2017}
K.T. Sch{\"u}tt, F.~Arbabzadah, K.R.~M{\"u}ller S.~Chmiela, and A.~Tkatchenko.
\newblock Quantum-chemical insights from deep tensor neural networks.
\newblock \emph{Nature Communications}, 8:\penalty0 13890 EP --, 01 2017.
\newblock URL \url{https://doi.org/10.1038/ncomms13890}.

\bibitem[Silver et~al.(2018)Silver, Hubert, Schrittwieser, Antonoglou, Lai,
  Guez, , Lanctot, Sifre, Kumaran, Graepel, Lillicrap, Simonyan, and
  Hassabis]{AlphaZero}
D.~Silver, T.~Hubert, J.~Schrittwieser, I.~Antonoglou, M.~Lai, A.~Guez, ,
  M.~Lanctot, L.~Sifre, D.~Kumaran, T.~Graepel, T.~Lillicrap, K.~Simonyan, and
  D.~Hassabis.
\newblock A general reinforcement learning algorithm that masters chess, shogi,
  and go through self-play.
\newblock \emph{Science}, 362\penalty0 (6419):\penalty0 1140--1144, 2018.
\newblock ISSN 0036-8075.
\newblock \doi{10.1126/science.aar6404}.
\newblock URL \url{https://science.sciencemag.org/content/362/6419/1140}.

\bibitem[Singh et~al.(2015)Singh, De, Zhang, Goldstein, and
  Taylor]{SiDeZhGoTa2015}
B.~Singh, S.~De, Y.~Zhang, T.~Goldstein, and G.~Taylor.
\newblock Layer-specific adaptive learning rates for deep networks.
\newblock \emph{ICMLA}, 2015.

\bibitem[Tibshirani(1996)]{LASSO1}
R.~Tibshirani.
\newblock Regression shrinkage and selection via the lasso.
\newblock \emph{Journal of the Royal Statistical Society. Series B},
  58\penalty0 (1):\penalty0 267--288, 1996.

\bibitem[Tieleman and Hinton(2012)]{rmsprop}
T.~Tieleman and G.~Hinton.
\newblock Lecture 6.5 - {R}{M}{S}prop: Divide the gradient by a running average
  of its recent magnitude.
\newblock \emph{COURSERA: Neural Networks for Machine Learning}, 2012.

\bibitem[Welling and Teh(2011)]{WeTe2011}
M.~Welling and Y.W. Teh.
\newblock Bayesian learning via stochastic gradient {L}angevin dynamics.
\newblock In \emph{Proceedings of the 28th International Conference on Machine
  Learning (ICML-11)}, pages 681--688, 2011.

\bibitem[Williams(1995)]{LASSO2}
P.~Williams.
\newblock Bayesian regularization and pruning using a {L}aplace prior.
\newblock \emph{Neural Computation}, 7:\penalty0 117--143, 1995.

\bibitem[Wilson et~al.(2017)Wilson, Roelofs, Stern, Srebro, and
  Recht]{against_adam}
A.C. Wilson, R.~Roelofs, M.~Stern, N.~Srebro, and B.~Recht.
\newblock The marginal value of adaptive gradient methods in machine learning.
\newblock \emph{arXiv:1705.08292}, 2017.

\bibitem[Xu et~al.(2015)Xu, Wang, Chen, and Li]{Xu}
B.~Xu, N.~Wang, T.~Chen, and M.~Li.
\newblock Empirical evaluation of rectified activations in convolutional
  network.
\newblock \emph{CoRR, arXiv: 1505.00853}, 2015.

\bibitem[Zeiler(2012)]{Ze2012}
M.~Zeiler.
\newblock Adadelta: An adaptive learning rate method.
\newblock \emph{CoRR}, 2012.

\bibitem[Zhang et~al.(2017)Zhang, Bengio, Hardt, Recht, and
  Vinyals]{RethinkingGeneralization}
C.~Zhang, S.~Bengio, M.~Hardt, B.~Recht, and O.~Vinyals.
\newblock Understanding deep learning requires rethinking generalization.
\newblock \emph{International Conference on Learning Representations}, 2017.

\bibitem[Zwanzig(1988)]{Zw1988}
R.~Zwanzig.
\newblock Diffusion in a rough potential.
\newblock \emph{Proc. Natl. Acad. Sci. USA}, 87:\penalty0 2029--2030, 1988.

\end{thebibliography}

\end{document}